\documentclass[letterpaper]{article} 
\usepackage{aaai24}  

\usepackage{times}  
\usepackage{helvet}  
\usepackage{courier}  
\usepackage[hyphens]{url}  
\usepackage{graphicx} 
\usepackage{natbib}  
\usepackage{caption} 
\usepackage{algorithm}
\usepackage{algorithmic}
\usepackage{svg}
\usepackage{newfloat}
\usepackage{listings}
\DeclareCaptionStyle{ruled}{labelfont=normalfont,labelsep=colon,strut=off} 
\floatstyle{ruled}
\newfloat{listing}{tb}{lst}{}
\floatname{listing}{Listing}
\usepackage{makecell}

\usepackage[utf8]{inputenc} 
\usepackage[T1]{fontenc}    
\usepackage{url}            
\usepackage{booktabs}       
\usepackage{amsfonts}       
\usepackage{nicefrac}       
\usepackage{microtype}      
\usepackage{xcolor}         

\usepackage{tikz}
\usepackage[utf8]{inputenc} 
\usepackage[T1]{fontenc}    
\usepackage{algorithm}
\usepackage{algorithmic}
\usepackage{lscape}
\usepackage[algo2e]{algorithm2e} 
\usepackage{url}            
\usepackage{booktabs}       
\usepackage{amsfonts}       
\usepackage{nicefrac}       
\usepackage{microtype}      
\usepackage{xcolor}         
\usepackage{times}
\usepackage{epsfig}
\usepackage{graphicx}
\usepackage{amsmath}
\usepackage{numprint}
\usepackage{amssymb}
\usepackage{booktabs}
\usepackage{rotating}
\usepackage{makecell}
\usepackage{amssymb}
\usepackage{pifont}
\newcommand{\xmark}{\ding{55}}%
\newcommand{\sparsemat}{\ensuremath{\gls{WeightMatrix}^{\mathrm{sp}}}}
\newcommand{\dependence}{\ensuremath{\gamma}}

\newcommand{\fgvcheader}{\gls{fgvcheader}}
\newcommand{\stanfordheader}{\gls{stanfordheader}}
\newcommand{\cubheader}{\gls{cubheader}}
\newcommand{\travelingheader}{\gls{travelingheader}}
\newcommand{\imgnetheader}{\gls{imgnetheader}}

\newcommand{\fheader}{A}
\newcommand{\sheader}{S}
\newcommand{\cheader}{C}
\newcommand{\theader}{T}
\newcommand{\iheader}{I}

\newcommand{\inpercent}{ in percent}

\usepackage{multirow}
\usepackage{multicol}
\usepackage{smartdiagram}
\usepackage{tikz}
\usepackage{csquotes}
\usepackage{caption}
\usepackage{arydshln}

\usepackage{subcaption}
\usetikzlibrary{shapes.geometric, arrows, fit}

\usepackage[acronym,nogroupskip]{glossaries}
\glsdisablehyper

\newglossaryentry{customLoss}
{
name={\ensuremath{\mathcal{L}_{\mathrm{div}}}},
description={Custom Loss für unterschiedliche Features}
}
\newglossaryentry{cLW}
{
name={\ensuremath{\beta}},
description={Gewichtung für customLoss}
}
\newglossaryentry{elaWeight}
{
name={\ensuremath{\lambda}},
description={Gewichtung für elasticNet}
}
\newglossaryentry{elaW}
{
name={\ensuremath{\alpha}},
long = {\ensuremath{\alpha\in[0,1]}},
description={Gewichtung zwischen l1 und l2 für elasticNet}
}
\newglossaryentry{trainDataset}
{
name={\ensuremath{\boldsymbol{D}_t}},
first = {\ensuremath{\boldsymbol{D}_t\in \mathbb{R}^{\gls{nTrainImages}\times 3\times w\times h}}},
description={LokalisierungMaps mit Missingness}
}
\newglossaryentry{nFeatures}
{
name={\ensuremath{n_{f}}},
description={Anzahl der verwendeten Features}
}
\newglossaryentry{nReducedFeatures}
{
name={\ensuremath{\gls{nFeatures}^*}},
description={Anzahl der verwendeten Features im Sparse Decision Layer}
}
\newglossaryentry{outputVector}
{
name={\ensuremath{\boldsymbol{y}}},
long = {\ensuremath{\boldsymbol{y}\in \mathbb{R}^{\gls{nClasses}}}},
description={Finaler Ausgang des Netzes}
}
\newglossaryentry{features}
{
name={\ensuremath{\boldsymbol{f}}},
description={Aus Bild berechnete Features}
}
\newglossaryentry{LocalizationMaps}
{
name={\ensuremath{\boldsymbol{L}}},
long = {\ensuremath{\boldsymbol{L}_p}\in \mathbb{R}^{\gls{nReducedFeatures}\times \frac{w}{p}\times \frac{h}{p}}},
description={LokalisierungMaps mit Missingness}
}
\newglossaryentry{featuresMapwidth}
{
name={\ensuremath{ w_M}},
description={Weite der FeatureMap}
}
\newglossaryentry{featuresMapheigth}
{
name={\ensuremath{ h_M}},
description={Weite der FeatureMap}
}
\newglossaryentry{featureMaps}
{
first ={\ensuremath{\boldsymbol{M} \in \mathbb{R}^{\gls{nFeatures} \times \gls{featuresMapwidth}\times \gls{featuresMapheigth}}}},
name={\ensuremath{m}},
plural={\ensuremath{\boldsymbol{M}}},
description={Aus Bild berechnete Features}
}
\newglossaryentry{trainFeatures}
{
first ={\ensuremath{\boldsymbol{F}^{\mathrm{train}} \in \mathbb{R}^{\gls{nTrainImages} \times \gls{nFeatures}}}},
name={\ensuremath{\boldsymbol{F}^{\mathrm{train}}}},
plural = {\ensuremath{F^{\mathrm{train}}}},
description={Aus Bild berechnete Features}
}
\newglossaryentry{denseNet}
{
first = {DenseNet121~\citep{huang2017densely}},
name ={DenseNet121},
description={Final layer in the neural network}
}
\newglossaryentry{resNet}
{
first = {Resnet50~\citep{he2016deep}},
name ={Resnet50},
description={Final layer in the neural network}
}
\newglossaryentry{incv}
{
first = {Inception-v3~\citep{szegedy2016rethinking}},
name ={Inception-v3},
description={Final layer in the neural network}
}
\newglossaryentry{birdsheader}
{
first = {NABirds~\citep{7298658}},
name = {NABirds},
description={Final layer in the neural network}
}
\newglossaryentry{fgvcheader}
{
first = {FGVC-Aircraft~\citep{FGVCAircraft}},
name={FGVC-Aircraft},
description={Final layer in the neural network}
}
\newglossaryentry{stanfordheader}
{
first = {Stanford Cars~\citep{StanfordCars}},
name={Stanford Cars},
description={Final layer in the neural network}
}
\newglossaryentry{cubheader}
{
first = {CUB-2011~\citep{wah2011caltech}},
long = {CUB-2011~\citep{wah2011caltech}},
name={CUB-2011},
description={Final layer in the neural network}
}
\newglossaryentry{travelingheader}
{
first = {TravelingBirds~\citep{koh2020concept}},
long = {TravelingBirds~\citep{koh2020concept}},
name={TravelingBirds},
description={Final layer in the neural network}
}
\newglossaryentry{decisionLayer}
{
name={decision layer},
description={Final layer in the neural network}
}
\newglossaryentry{fittingLossTarget}
{
name={\ensuremath{\mathcal{L}_{\mathrm{target}}}},
description={Main goal of fitting}
}
\newglossaryentry{layerName}
{
name ={SLDD-Model},
first ={SLDD-Model~\citep{norrenbrocktake}},
description={The proposed benchmark}
}
\newglossaryentry{NewlayerName}
{
name ={Q-SENN},
description={The proposed benchmark}
}
\newglossaryentry{denseLayer}
{
name={{dense high-dimensional \gls{decisionLayer}}},
description={The layer that results from training}
}
\newglossaryentry{correlationMatrix}
{
name={\ensuremath{\boldsymbol{Q}}},
first = {\ensuremath{\boldsymbol{Q}\in \mathbb{R}^{\gls{nFeatures}\times\gls{nFeatures}}}},
long = {\ensuremath{q}},
description={Correlation Matrix}
}
\newglossaryentry{featureVector}
{
name={\ensuremath{f}},
long={\ensuremath{\boldsymbol{f}}},
first ={\ensuremath{\boldsymbol{f} \in \mathbb{R}^{\gls{nFeatures}}}},
description={The features of the dense alyer}
}
\newglossaryentry{RedfeatureVector}
{
name={\ensuremath{\boldsymbol{f^*}}},
first ={\ensuremath{\boldsymbol{f^*} \in \mathbb{R}^{\gls{nReducedFeatures}}}},
description={The selected feature Vector}
}
\newcommand{\suppl}{Suppl.~}
\newglossaryentry{OnlyInteractionVector}
{
name={\ensuremath{\boldsymbol{P}}},
long={\ensuremath{\boldsymbol{P} \in \mathbb{R}^{ \gls{nInteractions}}}},
description={The Interaction Vector}
}
\newglossaryentry{InteractionVector}
{
name={\ensuremath{\boldsymbol{f^*_{\phi}}}},
first ={\ensuremath{\boldsymbol{f^*_{\phi}} \in \mathbb{R}^{\gls{nReducedFeatures} + \gls{nInteractions}}}},
description={The Extended Interaction Vector}
}
\newglossaryentry{nInteractions}
{
name={\ensuremath{n_I}},
description={number of interaction term}
}
\newglossaryentry{dnn}
{
name={\ensuremath{f_\theta(x})},
description={Deep neural network}
}

\newglossaryentry{bias}
{
name={\ensuremath{\boldsymbol{b}}},
long ={\ensuremath{\boldsymbol{b} \in \mathbb{R}^{\gls{nClasses}}}},
description={The bias in the decison layer}
}
\newglossaryentry{classifyFunc}
{
name={\ensuremath{C}},
long = {\ensuremath{C}},
description={The classifier on the featuers}
}
\newglossaryentry{WeightMatrix}
{
name={\ensuremath{\boldsymbol{W}}},
long = {\ensuremath{\boldsymbol{W}\in \mathbb{R}^{\gls{nClasses}\times \gls{nReducedFeatures} }}},
plural = {\ensuremath{w}},
description={The Weight matrix in the decision layer}
}
\newglossaryentry{qWeightMatrix}
{
name={\ensuremath{\boldsymbol{W}^{Q}}},
first ={\ensuremath{\boldsymbol{W}^{Q}\in \{-\alpha,0,\alpha\}^{\gls{nClasses}\times \gls{nReducedFeatures} }}},
long = {\ensuremath{\boldsymbol{W}^{Q}\in \{-\alpha,0,\alpha\}^{\gls{nClasses}\times \gls{nReducedFeatures} }}},
plural = {\ensuremath{w}},
description={The Weight matrix in the decision layer}
}

\newglossaryentry{nClasses}
{
name={n_c},
description={Number of Classes}
}
\newglossaryentry{nTrainImages}
{
name={\ensuremath{n_T}},
description={Number of Train Images}
}
\newglossaryentry{nWeights}
{
name={\ensuremath{n_w}},
description={Number of Entries != 0 in \gls{WeightMatrix}}
}
\newglossaryentry{nperClass}
{
name={\ensuremath{n_{wc}}},
description={Number of Entries != 0 in \gls{WeightMatrix} per Class}
}
\newglossaryentry{interpTrans}
{
name={\ensuremath{\phi}},
description={Interpretable Transformation}
}
\newglossaryentry{quantizedValue}
{
name={\ensuremath{\alpha}},
description={Interpretable Transformation}
}
\newglossaryentry{targetVector}
{
name={\ensuremath{\hat{\boldsymbol{y}}}},
description={Target Vector in Training}
}
\newglossaryentry{glmsaga}
{
first = {\mbox{\textit{glm-saga}~\citep{wong2021leveraging}}},
long = {\mbox{\textit{glm-saga}~\citep{wong2021leveraging}}},
name={\mbox{\textit{glm-saga}}},
description={Target Vector in Training}
}
\newglossaryentry{cbm}
{
name={CBM},
first ={\textit{Concept Bottleneck Model}~(CBM)~\citep{koh2020concept}},
long = {CBM~\citep{koh2020concept}},
description={Target Vector in Training}
}
\newglossaryentry{ProtoPNet}
{
name={\textit{ProtoPNet}},
first ={\textit{ProtoPNet}~\citep{chen2019looks}},
description={Target Vector in Training}
}
\newglossaryentry{PIP-Net}
{
name={\textit{PIP-Net}},
first ={\textit{PIP-Net}~\citep{nauta2023pipnet}},
description={Target Vector in Training}
}
\newglossaryentry{ProtoPShare}
{
name={\textit{ProtoPShare}},
first ={\textit{ProtoPShare}~\citep{rymarczyk2021protopshare}},
description={Target Vector in Training}
}
\definecolor{myGreen}{RGB}{34, 139, 34}
\newglossaryentry{ProtoPool}
{
name={\textit{ProtoPool}},
first ={\textit{ProtoPool}~\citep{rymarczyk2022interpretable}},
long ={\textit{ProtoPool}~\citep{rymarczyk2022interpretable}},
description={Target Vector in Training}
}
\newglossaryentry{imgnetheader}
{
first = {ImageNet-1K~\citep{imagenet15russakovsky}},
long = {ImageNet-1K~\citep{imagenet15russakovsky}},
name={ImageNet-1K},
description={Final layer in the neural network}
}
\newglossaryentry{ProtoTree}
{
name={\textit{Prototree}},
first ={\textit{ProtoTree}~\citep{nauta2021neural}},
description={Target Vector in Training}
}
\newglossaryentry{ImageSample}
{
name={\ensuremath{\boldsymbol{I}}},
first ={\ensuremath{\boldsymbol{I} \in \mathbb{R}^{3\times w\times h}}},
description={The classifier on the featuers}
}

\newcommand{\posmetric}[1]{\ensuremath{\mathrm{pos}^{\mathrm{#1}}}}

\newcommand{\incv}{\gls{incv}}
\newcommand{\resnet}{\glsname{resNet}}
\newcommand{\densenet}{\gls{denseNet}}

\newcommand{\attributeset}[1]{\ensuremath{\rho_{#1}}}

\newcommand{\loc}[1]{\textrm{diversity@#1}}
\newcommand{\eg}{\textit{e}.\,\textit{g}.}
\newcommand{\glm}{\gls{glmsaga}}
\newcommand{\glmtable}{glm-saga\textsubscript{5}} 
\newcommand{\densetable}{Dense} 
\newcommand{\tablefinisher}[1]{ dependent on #1}

\newcommand{\st}{\textit{s}.\,\textit{t}.}
\newcommand{\clipembedding}[1]{\ensuremath{\boldsymbol{#1}_{\mathrm{clip}}}}
\newcommand{\nclipfeatures}{\ensuremath{n_f^{\mathrm{clip}}}}
\newcommand{\gtmatrix}{\ensuremath{A}^{\mathrm{gt}}}
\newcommand{\gtvecvtor}{\ensuremath{\mathbf{a}^{\mathrm{gt}}}}
\newcommand{\clipmatrix}{\ensuremath{A}^{\mathrm{clip}}}
\newcommand{\clipvector}{\ensuremath{\mathbf{a}^{\mathrm{clip}}}}
\newcommand{\ourtable}{\gls{NewlayerName} (Ours)}
\newcommand{\fidelity}{\textbf{Fidelity}}
\newcommand{\diversity}{\textbf{Diversity}}
\newcommand{\grounding}{\textbf{Grounding}}
\newcommand{\supplm}{supplementary material}

\newcommand{\arrowDown}{\ensuremath{\boldsymbol{\downarrow}}}

\newcommand{\sresnet}{Resnet}
\newcommand{\sdensenet}{DenseNet}
\newcommand{\sinception}{Inception}
\newcommand{\slddtable}{\gls{layerName}~(baseline)}

\newcommand{\qca}{85.9} 
\newcommand{\qaa}{92.1} 
\newcommand{\qsa}{92.9} 
\newcommand{\qta}{67.3} 

\newcommand{\qcd}{78.1} 
\newcommand{\qad}{88.7} 
\newcommand{\qsd}{81.6}
\newcommand{\qtd}{75.4} 

\newcommand{\qcal}{2.2}
\newcommand{\qtal}{2.1} 

\newcommand{\qct}{30.0} 
\newcommand{\qat}{30.5} 
\newcommand{\qst}{29.3} 
\newcommand{\qtt}{30.7} 

\newcommand{\sca}{85.7}
\newcommand{\saa}{92.0} 
\newcommand{\ssa}{92.9}
\newcommand{\sta}{64.1} 

\newcommand{\scd}{67.2} 
\newcommand{\sad}{75.6} 
\newcommand{\ssd}{68.4} 
\newcommand{\std}{65.8} 

\newcommand{\scal}{1.9}
\newcommand{\stal}{1.7} 

\newcommand{\sct}{46.9} 
\newcommand{\sat}{53.1} 
\newcommand{\sst}{45.9} 
\newcommand{\stt}{46.7} 

\newcommand{\abbrv}{\cheader{}, \fheader, \sheader{}, \theader{} and \iheader{} abbreviate \glsname{cubheader}, \glsname{fgvcheader},  \glsname{stanfordheader}, \glsname{travelingheader} and \glsname{imgnetheader}. }
\newcommand{\arrowUp}{\ensuremath{\boldsymbol{\uparrow}}} 

\newcommand{\leftend}{-.27cm}

\newcommand{\comparison}[1]{
\begin{figure*}
     \begin{subfigure}[t]{.35\textwidth}
         \centering
         \includegraphics[width=\linewidth]{plots/SupplComparison/#1_Dense_5Weights.png}
        \caption{Conventional Dense \resnet{}}
     \end{subfigure}
     \hfill
     \begin{subfigure}[t]{.35\textwidth}
         \centering
         \includegraphics[width=\linewidth]{plots/SupplComparison/#1_Finetuned_5Weights.png}
            \caption{\ourtable{} with \resnet{}}
     \end{subfigure}
     \caption{Feature maps of the top 5 features by magnitude for class \detokenize{#1} on example images. The used weights for the respective features are also displayed. For~\gls{NewlayerName} all nonzero weights have the same magnitude $\alpha$. }
        \label{app:fig:vizExamples#1}
\end{figure*}
}
\tikzstyle{process} = [rectangle, minimum width=1cm, minimum height=1cm, text centered, text width=1.8cm, draw=black]
\tikzstyle{decision} = [diamond, minimum width=1cm, minimum height=1cm, text centered, draw=black]
\tikzstyle{arrow} = [thick,->,>=stealth]

\title{Q-SENN: Quantized Self-Explaining Neural Networks}
\author {
    Thomas Norrenbrock,
    Marco Rudolph,
    Bodo Rosenhahn
}
\affiliations {
    Institute for Information Processing (tnt)\\
  L3S - Leibniz Universität Hannover, Germany \\
   \{norrenbr, rudolph, rosenhahn\}@tnt.uni-hannover.de}

\usepackage{bibentry}

\begin{document}

\maketitle

\begin{abstract}

Explanations in Computer Vision are often desired, but most  Deep Neural Networks can only provide saliency maps with questionable faithfulness.
Self-Explaining Neural Networks (SENN) extract interpretable concepts with fidelity, diversity, and grounding to combine them linearly for decision-making. While they can explain \textit{what} was recognized, initial realizations lack accuracy and general applicability.  
We propose the Quantized-Self-Explaining Neural Network~\enquote{\gls{NewlayerName}}.
\gls{NewlayerName} satisfies or exceeds the desiderata of SENN while being applicable to more complex datasets and maintaining 
most or all
of the accuracy of an uninterpretable baseline model, outperforming previous work in all considered metrics. 
\gls{NewlayerName} describes the relationship between every class and  feature as either positive, negative or neutral  instead of an arbitrary number of possible relations, enforcing more binary human-friendly features. 
Since every class is assigned just $5$ interpretable features on average, \gls{NewlayerName} shows convincing local and global interpretability.
Additionally, we propose a feature alignment method, capable of aligning learned features with human language-based concepts without additional supervision. Thus, what is learned can be more easily verbalized. The code is published:  \url{https://github.com/ThomasNorr/Q-SENN}
\end{abstract}

 \section{Introduction}
 \begin{figure}[h!]
 
    \begin{subfigure}[t]{\linewidth}
    \centering
    \begin{tikzpicture}
      \node (goingback) at (.68\textwidth, 0)[ xshift=0.7cm, yshift=4.7cm] {\includegraphics[width=.25\textwidth]{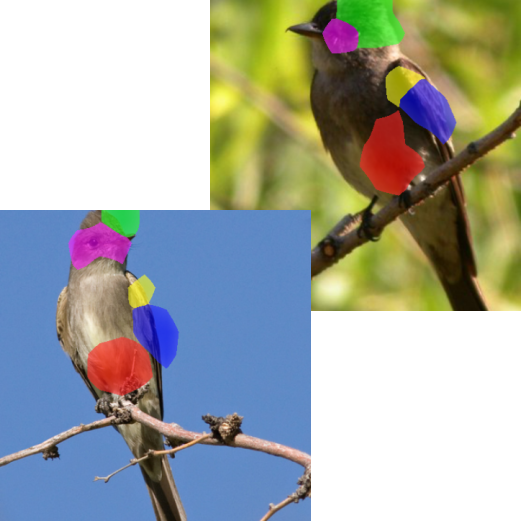}};
         \node[anchor=south west] (image) at (0,0) {\includegraphics[width=.65\textwidth]{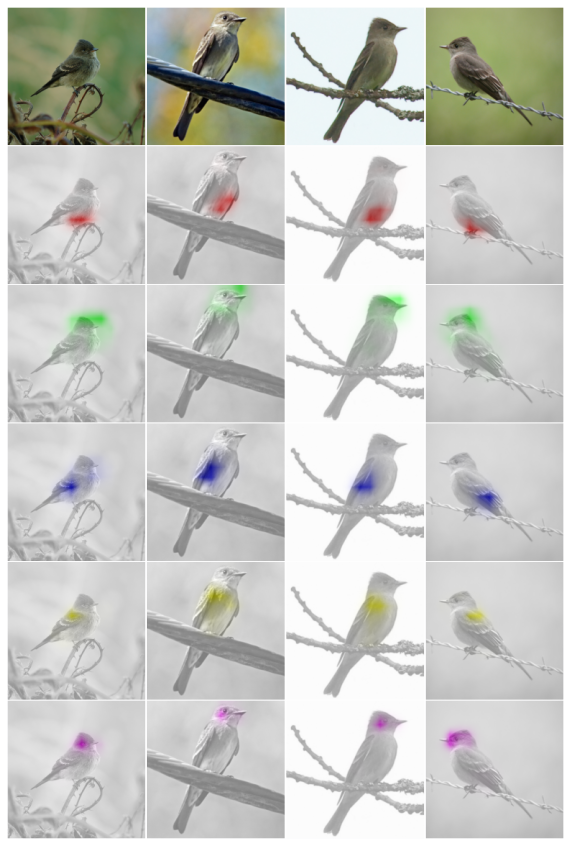}};
     \node (leftfidelitystart) at (image.south west){};
   \node (leftfidelityend) at (image.south west)[xshift=-.12cm, yshift=-.12cm]{};

       \draw[->] (leftfidelitystart) -- (leftfidelityend);
               \node (fidelitystart) at (image.south west) [xshift=.65\textwidth , yshift=.65\textwidth ] {};

            \node (fidelityend) at (image.south west) [xshift=7cm, yshift=7cm] {};
                 \draw[->] (fidelitystart) -- (fidelityend) node[midway,right=.1cm, above,  rotate=45] {\fidelity{}{}} coordinate[midway] (fidelitylabel);
            \node[font=\small, below] at (fidelitylabel)[ rotate=45] {Generalization};

             \node (groundingstart) at (image.south west) [xshift=-0cm, yshift=-.2cm] {};

            \node (groundingend) at (image.south east) [xshift=0cm, yshift=-.2cm] {};
                         \draw[<->] (groundingstart) -- (groundingend) node[midway,below] {\grounding{}};
      
              \node[font=\small, above] at (image.south)[yshift=-.3cm] {consistently localized on human attribute};

                        \node (diversitystart) at (image.south west) [xshift=\leftend, yshift=0cm] {};

            \node (diversityend) at (image.north west) [xshift=\leftend, yshift=-1.6cm] {}; \draw[<->] (diversitystart) -- (diversityend) node[midway,above=.1cm, rotate=90] {\diversity{}} coordinate[midway] (diversitylabel);
               \node[font=\small, below] at (diversitylabel)[ rotate=90] {Non-overlapping Activations};

    \end{tikzpicture}
    \label{fig:GlobalCover}
    \end{subfigure}
        \caption{
    \gls{NewlayerName} optimizes for Diversity, Grounding and Fidelity: The global explanation shows one example class being recognized through $5$ interpretable features that show high Diversity and Grounding, consistently localizing on the same meaningful human attributes across images, \eg{}, \textit{belly, crown, upper tail, upper wing and eye.}
    When measuring Fidelity, the features generalize to unseen data and the local explanation fits the class explanation. 
    Visualization techniques are based on overlaying color-coded feature maps,  described in the \supplm. }
\label{fig:Cover}
\end{figure}  

The ability to comprehend the decision-making process of deep learning models is gaining significance, especially for safety-critical applications like autonomous driving or medical diagnosis, where practitioners and legal requirements necessitate a thorough understanding of the decision and its reasoning~\citep{molnar2020interpretable, article}. However, the high dimensionality of image data has made it challenging to develop interpretable models for computer vision, with many problems still unsolved.
\citet{alvarez2018towards} proposed \textit{Self-Explaining Neural Networks} (SENN) which can be described as linear combination of interpretable concepts extracted from the input, but their realization lacked accuracy and applicability to complex datasets beyond MNIST.
They had three desiderata for the concepts: 
\fidelity{} refers to relevancy, \diversity{} to non-overlapping concepts, and \grounding{} to the alignment with a human concept.
We propose the \textit{Quantized-SENN} (Q-SENN), which satisfies these requirements.
\gls{NewlayerName}
is motivated by
former works such as the \gls{layerName}, \gls{PIP-Net},
 \gls{ProtoPool} or the \gls{cbm}, as they all aim to obtain human-understandable concepts as input to an interpretable classifier \gls{classifyFunc}. 
 Our proposed \gls{NewlayerName} improves the interpretability
 of \gls{classifyFunc} compared to all competitors as it is extremely sparse, using just very few, usually $5$ to be easy for all humans to follow~\cite{miller1956magical}, features to recognize a class, and ternary.
 That way, the relationship between a feature and a class is either a  positive ($+1$), negative ($-1$) or neutral (0) assignment with no sublevels.
For example, a dog might be positively related to the \textit{four-legged} feature, neutral to most features, such as color,
and negatively related to features that might be required to differentiate it from other classes in the dataset, \eg{}, \textit{feline} if multiple classes of cats are present.
Additionally, the features used for~\gls{classifyFunc} exhibit improved desiderata of SENN, thus making~\gls{NewlayerName} 
both globally and locally
interpretable, as shown in Figure~\ref{fig:Cover}.
Finally, \gls{NewlayerName} is more broadly applicable, as it does not need additional annotations and scales to ImageNet~\citep{imagenet15russakovsky}. 
 We create our~\gls{NewlayerName} through an iterative process of calculating the sparse ternary layer and fine-tuning the model with these fixed feature-class assignments.
 We select a sufficiently low number of features to ensure a sharing between the classes. 
 This leads to the emergence of more grounded, general features during one iteration and subsequent removal of an assignment based on spurious correlations on the next iteration.
 That way, the model converges to more robust assignments between grounded features and classes and shows exceptional robustness to spurious correlations. 
 As the ternary structure prohibits features from differentiating between two assigned classes, they naturally become more diverse, general and binary leading to improved
 explanations~\cite{lipton1990contrastive} and generalization on all investigated datasets.
The social science survey by \citet{miller2019explanation} suggests classifying the explanations of \gls{NewlayerName} as human-friendly due to them being contrastive, concise and general.
Additionally, we introduce metrics to estimate the \grounding{} of learned features as human concepts and distinguish them from class detectors.
Overall, we demonstrate that the proposed~\gls{NewlayerName} is an interpretable model where every class is assigned on average $5$ diverse and more grounded features as depicted in Figure~\ref{fig:Cover} with an exceptional robustness to spurious correlations. 
Finally, 
while the features of \gls{NewlayerName} are more alignable with any human concept, thus show \grounding{},
the manual alignment of learned features with human concepts is laborious.
To facilitate this process,
we demonstrate a method to automatically align the features of our proposed~\gls{NewlayerName} with human interpretable concepts using CLIP~\cite{radford2021learning} without requiring additional annotations and validate the method using the attributes contained in~\cubheader.

Our main \textbf{contributions} are as follows:
\begin{itemize}
\item We propose quantization for sparse decision layers in an iterative fine-tuning loop, leading to an Quantized-Self-Explaining Neural Network (\gls{NewlayerName}) which is easy to interpret for humans.
\item We measure the  self-explaining quality via \fidelity{}, \diversity{} and \grounding{} and show significant improvements on these metrics compared to previous work.

\item We demonstrate the high interpretability, increased accuracy and exceptional robustness to spurious correlations of our \gls{NewlayerName} on several benchmark datasets and architectures for image classification.
\item We propose a method for automatically aligning the features with human interpretable concepts without additional annotations using CLIP.
\end{itemize}
\glsresetall
\section{Related Work}
\label{sec:Related}
\begin{figure*}
\centering
\begin{tikzpicture}[node distance=3.4cm]
\node (denseTraining) [process] {Train Dense Model with \gls{customLoss}};
\node (featureSel) [process, right of=denseTraining] {Feature Selection};
\node (fitSparse) [process, right of=featureSel] {Calculate Sparse Quantized Layer};
\node (finetune) [process, right of=fitSparse] {Fine-tune Features};
\draw[dotted,rounded corners=5mm] ([shift={(-.4cm,.4cm)}]fitSparse.north west) rectangle node[below=1cm, right = .1cm]{$N$ Iterations} ([shift={(.4cm,-.6cm)}]finetune.south east);
\draw [arrow] (denseTraining) -- (featureSel);
\draw [arrow] (featureSel) -- (fitSparse);
\draw [arrow] (finetune.north) -- ++(0,0.2cm) -- ++(-2.4cm,0);
\draw [arrow] (fitSparse.east) -- (finetune.west) ;
\node(end) [process, right of=finetune] {Feature Alignment using CLIP};
\draw [arrow] (finetune) -- (end) node[midway, above] {\gls{NewlayerName}};
\end{tikzpicture}
\caption{Overview of our proposed pipeline to construct a \gls{NewlayerName}.} 
\label{fig:OverviewAppraoch}
\end{figure*}
Interpretable machine learning refers to models that are interpretable by design, as well as post-hoc methods that attempt to explain what a model has learned. Interpretability can be categorized as either local or global, referring to the interpretability of a single decision or the entire model, respectively~\citep{molnar2020interpretable}.

Research towards global post-hoc interpretability attempts to align learned representations with human-understandable concepts~\citep{kim2018interpretability,bau2017network,AlphaZero,Fel_2023_CVPR,yuksekgonul2022posthoc,zhang2018interpreting}.
While they usually rely on  auxiliary data, 
we introduce a new such method for our proposed model with no need for additional annotations. 
Saliency maps are local explanations showing which part of an input image is relevant to a prediction. 
While post-hoc methods, \eg{} Grad-CAM~\citep{selvaraju2017grad}, often lack properties such as shift invariance and faithfulness~\citep{kindermans2019reliability,adebayo2018sanity}, interpretable models with built-in saliency maps~\cite{bohle2022b,bohle2023holistically,stalder2022what, zhang2018interpretable} provide more faithful localizations.
Interpretable models are becoming increasingly relevant, since desired properties, that additionally improve the effectiveness of post-hoc techniques such as ours,  can be built-in.
For example, \gls{NewlayerName}'s  low number of total features and their easy-to-interpret sparse ternary assignment increases the value of aligning a single feature with a human concept using our proposed alignment method.



Another type of interpretable neural network is the \textit{Self-Explaining Neural Network}~\cite{alvarez2018towards} (SENN). Its most simple form is described as
\begin{equation}
    y(\boldsymbol{x}) = \gls{classifyFunc}(\boldsymbol{x})^Tf(\boldsymbol{x}),
\end{equation}
where $\boldsymbol{x}$ refers to the input, $f(\boldsymbol{x})$ to concepts extracted from the input and $\gls{classifyFunc}(\boldsymbol{x})$ to class specific weights for the concepts which should be independent of small changes to $f(\boldsymbol{x})$. 
Additionally, SENN postulates three desiderata for the concepts: \fidelity{} refers to the preservation of relevant information about the input. \diversity{} describes the need for non-overlapping concepts. Finally, \grounding{} refers to an alignment of the concept with a human interpretable one. We consider our model as Self-Explaining Neural Network with constant, quantized and sparse $C(\boldsymbol{x})$.
This exceeds the desired stability of $C(\boldsymbol{x})$ and makes the computationally expensive optimization obsolete, enabling the application to complex datasets. 

Alternatively, models like \gls{ProtoTree}, \gls{ProtoPNet}, \gls{ProtoPShare}, \gls{ProtoPool} and \gls{PIP-Net} use a deep feature extractor to learn prototypes from data. 
The similarities to these prototypes are then fed into interpretable models.
Their interpretability is however unclear, as  \citet{kim2021hive} and \citet{hoffmann2021looks} showed a discrepancy between the human and computed similarities.
Our proposed model is to some extent similar to \gls{ProtoPool} or \gls{PIP-Net}, as it also shares a small set of total features between all classes and only uses very few per class. 
Instead of prototypes, our ~\gls{NewlayerName} uses fewer features in total and per class, while showing increased accuracy.
The \Gls{cbm} initially predicts the annotated concepts in a given dataset, and subsequently employs a basic model to forecast the target category from the identified concepts.
This notion has undergone further investigation and expansion~\citep{sawada2022concept,zarlenga2022concept}.
Both \citet{margeloiu2021concept} and~\citet{elude} suggest that end-to-end training of the \gls{cbm} can result in the encoding of additional information beyond the targeted concepts, leading to a reduction in interpretability.
\citet{marconato2022glancenets} address this issue by disentangling the features and increase alignment. 
Our proposed method differs from \gls{cbm}, as it does not use supplementary annotations and generates a decision function \gls{classifyFunc} that is considerably more interpretable, as it is sparse and quantized. 
Finally, \citet{liu2023ternary} showed that ternary networks can be used to compress models while maintaining high accuracy. 
For high levels of sparsity, \citet{glandorf2023hypersparse} demonstrate a similar result.
Instead of compression,
\gls{NewlayerName} offers interpretability through its ternary and sparse final layer.
While measuring the interpretability of deep neural networks is an open task, increased interpretability is typically measured by reducing model complexity, such as the number of operations~\citep{yang2017scalable,friedler2019assessing,ruping2006learning} or the number of features~\citep{ruping2006learning}.
This motivated recent work on interpretable machine learning~\cite{nauta2023pipnet,Bod2023}
and the~\gls{layerName}.
\subsection{SLDD-Model}
The 
\textit{Sparse-Low-Dimensional-Decision-Model} (\gls{layerName}), 
uses on average $\gls{nperClass}=5$ features out of $\gls{nReducedFeatures}=50$ total features in the final layer to recognize a single class, since humans can follow a decision based on five cognitive aspects~\citep{miller1956magical}. 
After training a model with the \textit{Feature Diversity Loss}~\gls{customLoss}, 
which penalizes highly activated and weighted features that localize on the same region,
a subset of features
is selected
in a supervised manner and
\glm{}
is used
to compute the regularization path, from which they chose a solution with on average $5$ nonzero weights per class.
The model is then fine-tuned with the final layer fixed to this solution, \st{} the features adapt to it. 
The resulting \glsname{layerName} shows competitive accuracy while being more interpretable.
We use the~\glsname{layerName} as baseline for our experiments and demonstrate that our method increases all desiderata of a SENN by quantizing the sparse matrix and iteratively optimizing it.
\renewcommand{\fidelity}{Fidelity}
\renewcommand{\diversity}{Diversity}
\renewcommand{\grounding}{Grounding}
\section{Method}
We evaluate our proposed \gls{NewlayerName} in
image classification. The task involves the classification of an image \gls{ImageSample} of dimensions $w$ and $h$ into a single class $c \in \{c_1,c_2, \dots, c_{\gls{nClasses}}\}$ using a deep neural network $\Phi$ as feature extractor and classifier network \gls{classifyFunc}. The network $\Phi$ extracts feature maps \Gls{featureMaps} and aggregates them into a feature vector \gls{featureVector}. The final output is obtained by applying \gls{classifyFunc} to the feature vector, resulting in the output vector \glsentrylong{outputVector} as $ \gls{outputVector} = \gls{classifyFunc}(\glsentrylong{featureVector})$.

\subsection{\Gls{NewlayerName}}
The proposed pipeline to create a \gls{NewlayerName} is shown in Figure~\ref{fig:OverviewAppraoch}. We first train a dense model and then select~\gls{nReducedFeatures} out of the \gls{nFeatures}~features, following the SLDD-Model,  to compute a low-dimensional sparse quantized final layer for the selected features using~\glm{}.
The model is then fine-tuned with the final layer fixed to that computed solution, \st{} the features adapt to the assigned classes.
The cycle of computing 
an interpretable final layer
and fine-tuning the model is repeated  $N$ times. Afterwards, the learned features can be aligned with human concepts in form of natural language to create 
a model with verbalized explanations.
One such method using CLIP is presented at the end of this paper.

\glsname{NewlayerName} is considered 
as SENN with constant quantized $\gls{classifyFunc}(\glsentrylong{featureVector})= \gls{outputVector} = \glsname{qWeightMatrix} \glsentrylong{featureVector} + \gls{bias} = \alpha \boldsymbol{W}^1\glsentrylong{featureVector} + \gls{bias}$
with the ternary sparse weight matrix \gls{qWeightMatrix}, 
or matrix of ones $\boldsymbol{W}^1\in\{-1,0,1\}^{\gls{nClasses}\times \gls{nReducedFeatures}}$,   
and bias \glsentrylong{bias}. Therefore, we want the features \glsentrylong{featureVector} of our model to show \fidelity{}, \diversity{} and \grounding{}. In this work, \fidelity{} is measured as accuracy, as preserving relevant information is required to accurately predict the class. 
For \diversity{},  the Feature Diversity Loss~\gls{customLoss}~\cite{norrenbrocktake} is utilized
to ensure a high local diversity of the feature maps~\Glspl{featureMaps} that are used for the same class.
The training loss is
calculated as
\begin{align}
    \mathcal{L}_{\mathrm{final}} = \mathcal{L}_{CE} + \gls{cLW}\gls{customLoss}
\end{align}
with $\gls{cLW}\in\mathbb{R}_+$ as weighting factor between cross-entropy loss and~\gls{customLoss}.
Additionally,   \gls{NewlayerName} uses an average of just $\gls{nperClass}=5$ features per class, \st{} redundant features are prevented and humans can follow the decision.
For \grounding{}, a reduced feature vector~\gls{RedfeatureVector} is used with $\gls{nReducedFeatures}<\gls{nClasses}$ features, enforcing them to capture general attributes shared by multiple assigned classes. 
Additionally, we quantize the sparse weight matrix~$\sparsemat\in \mathbb{R}^{\gls{nClasses}\times \gls{nReducedFeatures}}$, \st{} it only contains either $0$ or $\pm\gls{quantizedValue}$ with $\gls{quantizedValue}\in\mathbb{R}$. This 
leads to 
increased local and global interpretability, as the relationship between feature and class can be described as a positive, neutral, or negative assignment. 
It also hinders features from becoming class detectors,
which hurts the interpretability as a concept.   
Quantized weights counteract this, as increasing the activation does not move the prediction towards one specific class.
Therefore, the model is biased to learn more binary concepts, which lead to contrastive human-friendly explanations.
Finally, we designed the pipeline to do $N$ iterations of calculating the sparse matrix~\gls{qWeightMatrix} and fine-tuning it.
The features become more general during the sparse and low-dimensional fine-tuning, as they have to recognize concepts used for multiple classes across the entire dataset.
Therefore,
assignments based on spurious correlations become suboptimal.
The iteration leads to the removal of such assignments, \st{} more robust assignments remain. 
This leads to more grounded and robust features, thus supporting \grounding{} and \fidelity{}.  
To facilitate the convergence of~\gls{qWeightMatrix}, the learning rate decays while iterating.
\subsubsection{Quantization}
\label{sec:Quantization}

\begin{table*}[h]

\resizebox{\textwidth}{!}{
\centering
\begin{tabular}{l||ccc||ccc||ccc|cccc}
\toprule 
\multirow{2}{*}{Method}&  \multicolumn{3}{c||}{ Accuracy  \arrowUp}& \multicolumn{3}{c||}{ \loc{5} \arrowUp} & \multicolumn{1}{c}{No Concept} &\multicolumn{2}{c|}{Alignment $r$ \arrowUp}& \multicolumn{1}{c}{ Sparse} & \multicolumn{3}{c}{ Dependence \dependence{} \arrowDown}\\
& \cheader{} & \theader{}& \iheader & \cheader{} &  \theader{} & \iheader &Supervision & \cheader{} & \theader{}&
$\gls{nperClass}\leq5$
&\cheader{}& \theader{} & \iheader \\  
\midrule
Dense & 86.7  & 39.4 & 76.1  & 51.6  &  51.5 & 52.0 & \checkmark& 1.0 & 1.0 & \xmark & 10.4 &  22.3 & 3.2   \\ 
\midrule
\glmtable{}  & 77.7 & 36.0 & 58.0 & 46.7& 47.7 & 50.0 & \checkmark& 1.0 & 1.0 & \checkmark & 49.2 & 52.7 & 53.6 \\
\gls{cbm}-joint & 82.2 &36.9  & N/A & 71.0 & 73.7  & N/A &\xmark&  3.0 &   2.8 & \xmark&  6.1 &  5.7 &  N/A   \\
SLDD-Model & \sca  & \sta& 72.7 & \scd &\std & 58.8 & \checkmark& \scal  &  \stal& \checkmark & \sct &  \stt &   47.3\\  
\hdashline

Q-SENN (Ours) &\textbf{\qca}  & \textbf{\qta}& \textbf{74.3} & \textbf{\qcd}  & \textbf{\qtd} & \textbf{77.2} & \checkmark& \textbf{\qcal}  &  \textbf{\qtal}& \checkmark   & \textbf{\qct} &  \textbf{\qtt} &   \textbf{30.7}\\ 
\bottomrule
\end{tabular}
}
\caption{Comparison across all desired metrics with~\resnet{}. Best results among comparable interpretable models  are in bold.
\textit{N/A} indicates inapplicability due to missing annotations, 
while values with \xmark{} cannot be reasonably compared to the remaining rows due to sparsity or supervision.
Accuracy, \loc{5} and Dependence~\dependence{} are measured \inpercent{}. Note, that our proposed \gls{NewlayerName} maintains most or all of the accuracy of dense models while simultaneously heavily improving the interpretability.}
\label{tab:Comp}
\end{table*}

\begin{figure}
\begin{center}
  \includegraphics[width=\linewidth]{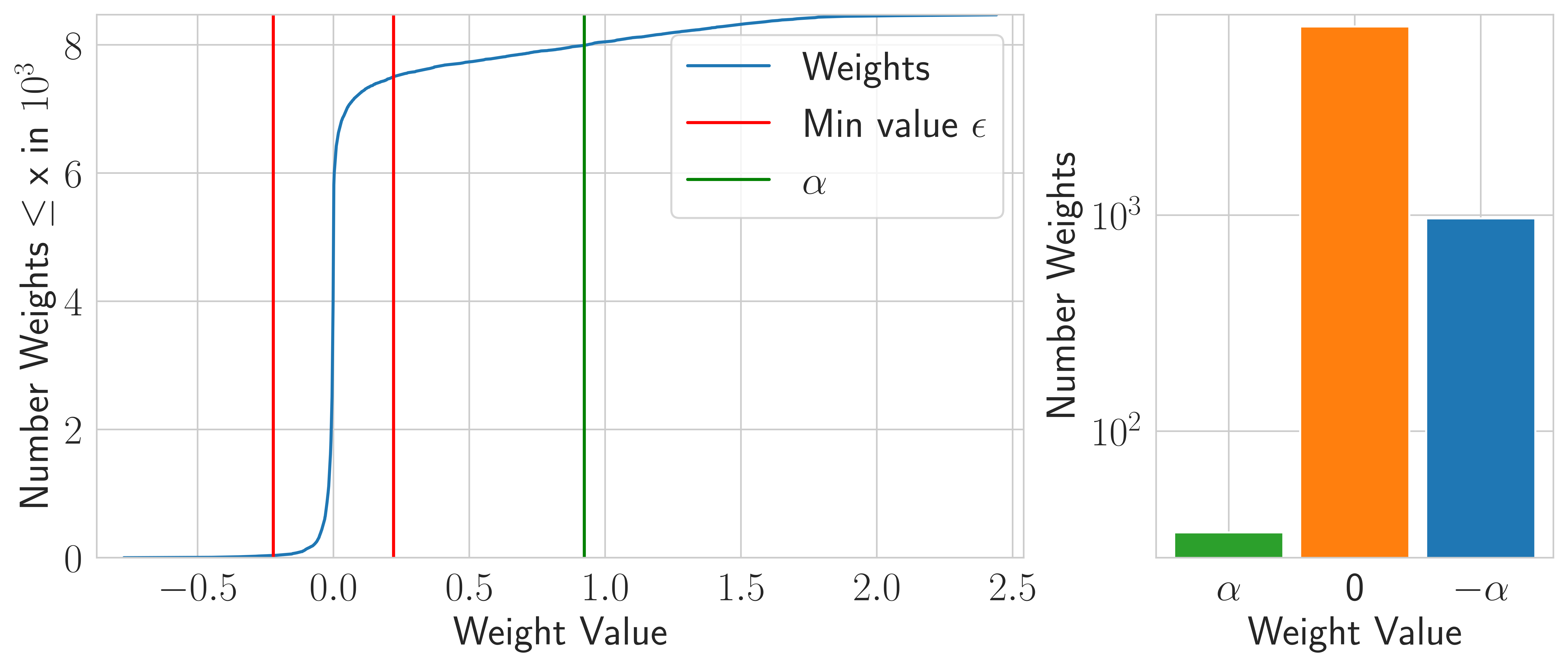}
\end{center}
    \caption{Exemplary result (right) of quantization on cumulative distribution (left) of nonzero weights in  \sparsemat{} for \glsname{cubheader}($\gls{nWeights}=1000)$: Weights are set to $0$ or $\pm\alpha$. 
  $\alpha$ is the average of all remaining values above $0$.}
  \label{fig:quantization}
 \end{figure}
For quantization, we start with the regularization path computed by~\glm{}. 
Figure~\ref{fig:quantization} gives an overview 
with
an exemplaric distribution of the nonzero weights in one of the sparse matrices \sparsemat{} before the first iteration.
The very small magnitude of most weights indicates overfitting on the training set, while only the entries with higher magnitude generalize~\cite{StateOfSparsity}.
To get more such entries in a sparse matrix, \sparsemat{} that resulted from \glm{} with the lowest regularization is used for quantization.
As meaningful assignments between features and classes are desired, we set a threshold $\epsilon$, below which, in terms of magnitude, all weights in \sparsemat{} are zeroed out. 
This threshold is calculated, \st{} exactly the desired number of weights \gls{nWeights}, in this paper usually $5*\gls{nClasses}$, remain:
\begin{equation}
       \epsilon =   \operatorname{sort}(\left|\sparsemat{}\right|)_{\gls{nWeights}}\quad.
       \label{eq:sort}
\end{equation}

In Equation~\ref{eq:sort}, $\left|\sparsemat{}\right|$ represents the matrix obtained by taking the absolute value of each element in $\mathbf{W}$. The expression $\operatorname{sort}(\left|\mathbf{W}\right|)$ describes the sorted list of absolute values of the elements in \sparsemat{} in descending order.
Afterwards all entries whose absolute value is~$\geq\epsilon$ are set to the average of the remaining values~$\alpha$. This leads to the quantized sparse Matrix~\glsentrylong{qWeightMatrix}:
\begin{align}
    \alpha &= \frac{\sum_{i,j} \mathbf{1}_{ \left|\sparsemat_{i,j}\right| \geq \epsilon} \left|\sparsemat_{i,j}\right|}{\sum_{i,j} \mathbf{1}_{ \left|\sparsemat_{i,j}\right| \geq \epsilon }}\\
    \gls{qWeightMatrix}_{i,j} &=
\begin{cases}
\alpha & \text{if } \sparsemat_{i,j} \geq \epsilon \\
-\alpha  & \text{if } \sparsemat_{i,j} \leq -\epsilon \\
0 & \text{otherwise}
\end{cases}\quad.
\end{align}
The indicator function $\mathbf{1}_{{ \left|\sparsemat_{i,j}\right| \geq \epsilon }}$ takes a value of 1 if $\left|\sparsemat_{i,j}\right|\geq\epsilon$, and 0 otherwise.
Overall, the quantization scheme ensures that the resulting~\gls{qWeightMatrix} contains the desired number of entries which can be meaningfully averaged to get assignments, as they all initially indicated a generalizing connection between feature and class.

\begin{table}
\begin{center}
\resizebox{\linewidth}{!}{
\centering
\begin{tabular}{c|c|c|c|c|c }
\toprule
Dataset & \cheader&\fheader & \sheader{}& \theader{}&      \iheader\\ 
\midrule
\# Classes $\gls{nClasses}$ & 200&100 &196 & 200 &   1000\\
\# Training & \numprint{5994}&\numprint{6667}  &\numprint{8144} &   \numprint{5994} &  \numprint{1281167}\\
\# Testing & \numprint{5774}&\numprint{3333} &\numprint{8041} &   \numprint{5774} &  \numprint{50000}\\
\bottomrule
\end{tabular}
}
\caption{Number of classes, training and testing samples of the used datasets. \abbrv }
\label{table:DatasetOverview}
\end{center}
\end{table}
\section{Experiments}
\label{sec:experiments}
\glsreset{resNet}
\glsreset{denseNet}
\glsreset{incv}
\glsreset{cubheader}
\glsreset{travelingheader}
\glsreset{stanfordheader}
\glsreset{fgvcheader}
\glsreset{imgnetheader}
This section presents the experimental results of our proposed method. The effectiveness  of our approach is evaluated using \gls{resNet}, \gls{denseNet}, and \gls{incv} as backbones.
We used the default datasets for fine-grained image classification and prototype-based methods, as well as \imgnetheader{} to showcase the applicability to large-scale datasets.
 For measuring the robustness against spurious correlations, \travelingheader{} is used. It is a dataset based on \cubheader{}, where the background is artificially spuriously correlated to the class in the training set. In the test set, the correlation is not maintained. Examples are in the \supplm{} and in Figure~\ref{fig:DivEx}.
Models that maintain their accuracy on this dataset are therefore less susceptible to spurious correlations.
An overview of the datasets, including \gls{cubheader} (in tables abbreviated C), \travelingheader{} (T), \gls{stanfordheader} (S), \fgvcheader{} (A) and the large-scale \imgnetheader{} (I) is provided in Table~\ref{table:DatasetOverview}.
Notably, \gls{cubheader} includes both attribute and class labels, enabling us to assess the alignment of learned features with semantically meaningful concepts.
Unless stated otherwise, reported values are averaged across five seeds. Implementation details and extensive results with standard deviations can be found in the \supplm. 
In all tables, \glmtable{} refers to applying \glm{} to a conventionally trained dense model with no feature selection and selecting a solution with $\gls{nperClass}\leq5$ and \textit{Dense} refers to a conventionally trained model without feature selection and a densely connected final layer.
For \gls{NewlayerName}, the number of features per class is set to $\gls{nperClass}=5$ and the number of total features to $\gls{nReducedFeatures}=50$.
Table~\ref{tab:Comp} gives an overview of all considered metrics.
We evaluate~\fidelity{} with accuracy on fine-grained and large-scale (I) image classification, as well as robustness to spurious correlation (T), \diversity{} with \loc{5} and \grounding{} with dependence~\dependence{} and alignment~$r$.
Notably, \gls{NewlayerName} maintains more than  $97\%$ of the dense model's accuracy  on \imgnetheader{}, while most interpretable competitors lack the capacity or annotations to function at that scale. 
Overall, \gls{NewlayerName} improves the baseline SLDD-Model on \fidelity{}, \diversity{} and \grounding{}, which are discussed
in detail 
in the following sections.

\begin{table*}
\centering
\begin{tabular}{c|ccc|ccc|ccc|ccc}
\toprule
&\multicolumn{3}{c|}{CUB-2011}&\multicolumn{3}{c|}{FGVC-Aircraft}&\multicolumn{3}{c|}{Stanford Cars}&\multicolumn{3}{c}{TravelingBirds}\\
Method & Inc. & Dense. & Res.   & Inc. & Dense. & Res.&   Inc. & Dense. & Res. &   Inc. & Dense. & Res. \\
\toprule
\densetable{} & 83.5&87.2&86.7   & 91.4& 92.5 &92.0 & 92.6 & 93.6 & 93.5  & 47.9 & 45.1 & 39.4\\  
\midrule
\glmtable{} & 76.9 & 69.8 & 77.7   & 89.1 & 86.2 & 89.9 & 88.9 & 81.9 & 89.1 & 45.7 & 42.7 & 36.0 \\ 
\gls{cbm}-joint & 80.1 & - & 82.2  & N/A & N/A&N/A   & N/A &  N/A& N/A& 51.8  & - & 36.9 \\
SLDD-Model & 80.8& 84.4&\sca   & 90.6& 92.1 &\saa  & 91.3 & 92.2 &\textbf{\ssa}  & \textbf{51.9} & 56.1 &\sta \\ 
\hdashline
Q-SENN (Ours) & \textbf{81.7} & \textbf{85.4} & \textbf{\qca}& \textbf{90.8} & \textbf{92.5} & \textbf{\qaa}  & \textbf{91.8} & \textbf{92.8} & \textbf{92.9} &   50.5 & \textbf{59.6} & \textbf{67.3} \\ 
\bottomrule

\end{tabular}
\caption{\fidelity{} measured as accuracy\tablefinisher{backbone}. 
Best result among more interpretable models are in bold.
Inc., Dense., and Res. abbreviate  \sinception , \sdensenet{} and \sresnet.
}
\label{table:Fidelity}
\end{table*}


\begin{table*}[h]
\resizebox{\textwidth}{!}{
\centering
\begin{tabular}{l||ccc||ccc||cc|ccc|ccc}
\toprule 
\multirow{2}{*}{Method}& \multicolumn{3}{c||}{ Accuracy  \arrowUp}& \multicolumn{3}{c||}{ \loc{5} \arrowUp} & \multicolumn{2}{c|}{Alignment $r$ \arrowUp} & \multicolumn{3}{c|}{ Dependence \dependence{} \arrowDown} & \multicolumn{3}{c}{Binary Features \arrowUp}\\
& \cheader{} &   \sheader & \theader{}&  \cheader{}  &   \sheader & \theader{} &  \cheader{} & \theader{}&
\cheader{} &   \sheader & \theader{} &\cheader{} &    \sheader & \theader{}  \\ 
\midrule
Q-SENN (Ours) &\textbf{\qca} &   \qsa & \qta&  \textbf{\qcd} &  \textbf{\qsd} & \textbf{\qtd} &  \qcal  &  \qtal  & \qct  &\qst & \qtt & 80.4 &  90.0 & \textbf{84.0}  \\ 
w/o Quantization & 85.4  &  92.6 & \textbf{70.0} & 64.4   &  65.3 & 61.8 & \textbf{3.1} &  \textbf{2.8} & 46.8 &  47.7 & 47.3 & 53.2 &  59.2 & 70.8  \\
w/o Iteration & \textbf{85.9} &  \textbf{93.0} & 63.9& 70.9&  73.5 &   68.9 & 1.7 & 1.5 & \textbf{29.2}& \textbf{28.6}  & \textbf{30.5 }& \textbf{90.4} & \textbf{96.4} & 78.4  \\ 
\bottomrule
\end{tabular}
}
\caption{Ablation Study on Impact of iteration and quantization. Binary Features are the percentage of features for which mean shift applied on the training distribution returns exactly 2 clusters.}
\label{tab:abl}
\end{table*}

\subsection{Fidelity}
\fidelity{} refers to the preservation of relevant information about the input in the concepts, which enables high accuracy.
Table~\ref{table:Fidelity} shows the impact of our method on accuracy with respect to backbones in detail.
On all conventional datasets, our proposed changes to the~\glsname{layerName} maintain or improve accuracy, while increasing interpretability.
 \gls{NewlayerName} also clearly surpasses prototype-based models, using their reduced image size of $224$, shown in Table~\ref{tab:proto-table}:
 The accuracy is increased with fewer features in total and per class, thus while improving interpretability.
 Compared to all competitors, our \gls{NewlayerName} shows exceptional robustness to spurious correlations, setting a new SOTA, while  maintaining $78\%$ of the accuracy of \glsname{cubheader} on \glsname{travelingheader} compared to the $45\%$ of the dense model.
 \gls{PIP-Net} and the \glsname{layerName} show that sparsely using grounded features during training, as opposed to \glmtable{} or \gls{ProtoPool}, is crucial for that robustness.
Considering the ablations in Table~\ref{tab:abl}, quantization seems to improve accuracy on conventional datasets, whereas the iteration leads to more robustness 
as spurious correlations get ignored 
with \gls{NewlayerName} combining the strengths.
We focus on~\resnet{} in the remaining part of the paper as the efficacy of \gls{NewlayerName} is apparent for all backbones. 

\begin{table}[htbp]
\resizebox{\linewidth}{!}{
{\fontsize{15pt}{17pt}\selectfont
  \centering
  \begin{tabular}{l|ccc|ccc|ccc}
    \toprule
    Method &  \multicolumn{3}{c|}{Accuracy \arrowUp} &
    \multicolumn{3}{c|}{Total Feat.\arrowDown} & \multicolumn{3}{c}{Feat./ Class\arrowDown} \\
   & C& S & T & C& S & T & C& S &T
    \\
    
    \midrule
    PIP-Net & 82.0 & 86.5 & 70.1  & 731 &669& 825  & 12 &11 & 5.6 \\
    ProtoPool & 85.5 &88.9& 42.9& 202 & 195 &202 &202 & 195 & 202\\
     \midrule
     Q-SENN & 84.7 & 91.5 &76.7& \textbf{50} & \textbf{50} &\textbf{50}& \textbf{5}& \textbf{5} &\textbf{5} \\
    $\gls{nReducedFeatures}>50$ & \textbf{86.4} & \textbf{92.2}& \textbf{78.4} & 202 &195& 202& \textbf{5} &\textbf{5}  &\textbf{5} \\
    \bottomrule
  \end{tabular}
  }
  }
  \caption{Comparison with state-of-the-art prototype-based methods with \resnet{} as backbone on the same data: With reduced sizes, \gls{NewlayerName} shows increased accuracy.}
  \label{tab:proto-table}
  
\end{table}

    
  
\subsection{Diversity}
\begin{figure}
\centering        \includegraphics[width=\linewidth]{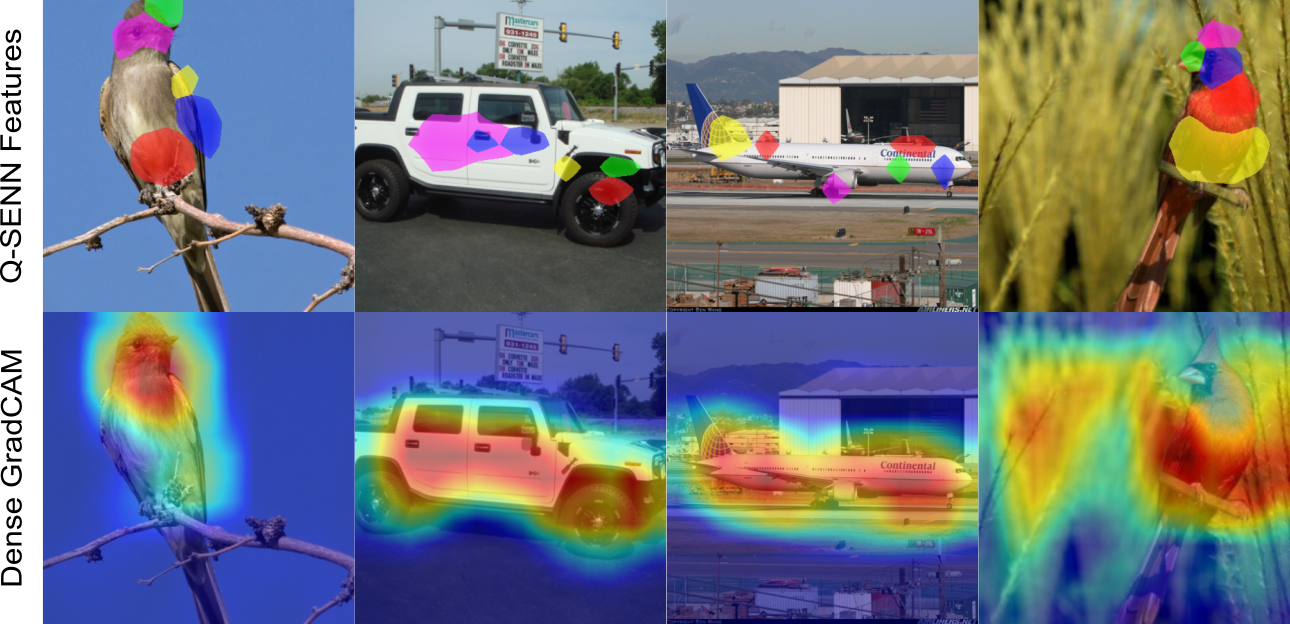}
        \caption{Exemplary local explanations
        in comparison:
        \gls{NewlayerName} offers explanations based on interpretable features.}
        \label{fig:DivEx}

\end{figure}
\diversity{} refers to the representation of inputs with non-overlapping concepts.
For measuring the \diversity{}, we utilize \loc{5}~\cite{norrenbrocktake} defined on 
$\boldsymbol{S}_5$ obtained by applying softmax to the $5$ highest weighted feature maps $\Glspl{featureMaps}_5$ for the predicted class as:
\begin{equation}
    \loc{5} = \frac{\sum_{i=1}^{\gls{featuresMapheigth}}\sum_{j=1}^{\gls{featuresMapwidth}}\max(S^1_{ij},S^2_{ij}, \dots, S^5_{ij})}{5}. 
\end{equation}
It measures the local diversity of the $5$ highest weighted features.
As shown in the \supplm{}, a higher \loc{5} of class-independent features also indicates less correlated features.
It is apparent in Table~\ref{tab:Comp} that the proposed~\gls{NewlayerName} shows the highest~\loc{5}, mainly pushed by quantization as the features can not recognize entire classes and further increased through iterations, shown in Table~\ref{tab:abl}.
Similar observations can be made on other architectures, provided in the \supplm.
The high~\loc{5} leads combined with the average number of just $5$ features per class to a globally more interpretable model. 
This is demonstrated in Figure~\ref{fig:Cover}:
The shown class is recognized through $5$ features that all consistently localize on different semantically meaningful regions, such as \textit{belly, crown, upper tail, upper wing and eye},  and are therefore easier to interpret and align. 
This makes feature alignment easier and also enables improved local explanations, as shown in Figure
~\ref{fig:DivEx}.
More examples with comparison to a dense model and global explanations are shown in  the \supplm.
\subsection{Grounding}
\label{sec:Grounding}
We propose two metrics to measure \grounding, the alignability with human concepts.
We evaluate the correlation between the features and attributes and their generality, as we aim for features that can be aligned with a human concept rather than an entire class.
To estimate this generality, the average dependence \dependence{} of the predicted class on the most important feature is measured. 
The dependence \dependence{} with the effect $e_i$ of every feature for the predicted class for every sample $i$ is calculated as: 
\begin{align}
    e_i &= \{\left|\gls{WeightMatrix}_{\hat{c},0} \cdot \gls{featureVector}_0\right|, \left|\gls{WeightMatrix}_{\hat{c},1} \cdot \gls{featureVector}_1\right|, \dots, \left|\gls{WeightMatrix}_{\hat{c},\gls{nFeatures}}  \cdot \gls{featureVector}_{\gls{nFeatures}}\right|\}\\
\dependence{} &= \frac{1}{\gls{nTrainImages}} \sum_{i=1}^{\gls{nTrainImages}} \frac{max(e_i)}{\sum_{j=1}^{\gls{nFeatures}}e_{ij}}\quad.
\end{align}
The dependence~$\dependence{}\in[0,1]$ measures how strongly the decision of a model usually depends on the most important feature and is designed to estimate how class-specific the features of a model are. 
The \textit{Pytorch} pretrained weights for \resnet{} on \imgnetheader{} validate this metric:
Using \glm{}, \textit{V2} shows $66\%$ accuracy with 1.1 features per class, while \textit{V1} shows $58\%$ with 4.7 features.
Thus, the features for \textit{V2} are already class detectors with no conceptual meaning.
This is reflected in the dependence with $44\%$ (i.\,e. $44\%$ of the prediction is on average based on one out of $2048$ features) for \textit{V2} compared to $3\%$ for \textit{V1}.
Because we aim for grounded features, we use \textit{V1} for our experiments.

Additionally, the correlation between the attributes contained in \cubheader{}, with $\attributeset{a+}$ denoting the indices where attribute $a$ is present and $\attributeset{a-}$ the opposite, and the features on the training set~\gls{trainFeatures}
\begin{align}
\gtmatrix_{a,j}&=
\frac{1}{\lvert\attributeset{a+}\rvert }\sum_{i\in\attributeset{a+}}\glspl{trainFeatures}_{i,j}- \frac{1}{\lvert\attributeset{a-}\rvert}\sum_{i\in\attributeset{a-}}\glspl{trainFeatures}_{i,j} \label{eq:Agt}
\end{align}
is measured, with the average maximum alignment $r$ per feature as comparable metric:
\begin{align}
    r = \frac{1}{\gls{nReducedFeatures}}\sum_{j=1}^{\gls{nReducedFeatures}} \frac{\gls{nTrainImages}}{\sum_{l=1}^{\gls{nTrainImages}}\glspl{trainFeatures}_{l,j}- \min_{l}\glspl{trainFeatures}_{l,j} }\max_{i} \gtmatrix_{i,j}\quad.
\end{align}
A higher value of $r$ indicates features that are more aligned with the attributes in~\cubheader{}.
We norm the difference by the average activation to be less dependent of assumptions about the underlying distribution.
Because the models shown are not trained to focus on the given attributes, it can be assumed that models with higher $r$ generally learn features more correlated to human concepts.
Table~\ref{tab:Comp} shows improvements in alignment $r$ and dependence \dependence{} across datasets.
Note that a dense matrix often leads to a low dependence~\dependence{} at the cost of global interpretability and \gls{cbm} was trained to predict a subset of the probed attributes. 
The impact of both proposed changes is apparent in Table~\ref{tab:abl}: The iteration increases the alignment with given attributes, as the features do not have to adapt to a sparse structure based on spurious correlations. 
The problem of high dependence~\dependence{} however, is reduced through the quantization.
In summary, only the proposed~\gls{NewlayerName} uses more aligned, less class-specific features. 
We additionally included the fraction of binary-like features.
A feature is considered as binary if the mean shift clustering, using a GPU-accelerated implementation by~\cite{SchRei2022a}, applied to the feature distribution on the training set returns exactly 2 clusters. 
As expected, the quantization leads to more binary features.
For reference, a Dense~\resnet{} on \cubheader{} has 15\% 
binary features and \gls{cbm} 67\%.
The clustering is visualized and analyzed in the \supplm.
The analysis shows that the few binary features of the Dense model, and only its features, are class-detectors instead of general features.
For \resnet, the Dense model has effectively no feature, one across 5 seeds, whose active cluster, the one with a higher mean, has more samples than the most frequently occuring training class.
In contrast, every binary feature of our \gls{NewlayerName} has this property, e.g. 80.4\% of all features on \gls{cubheader}. Notably, even the CBM, trained to predict binary attributes, has a few features that do not have this property.

\subsection{Interpretability Tradeoff}
\label{sec:Tradeoff}
\begin{figure}
\centering
     \begin{subfigure}[t]{.49\linewidth}
         \centering
         \includegraphics[width=\textwidth]{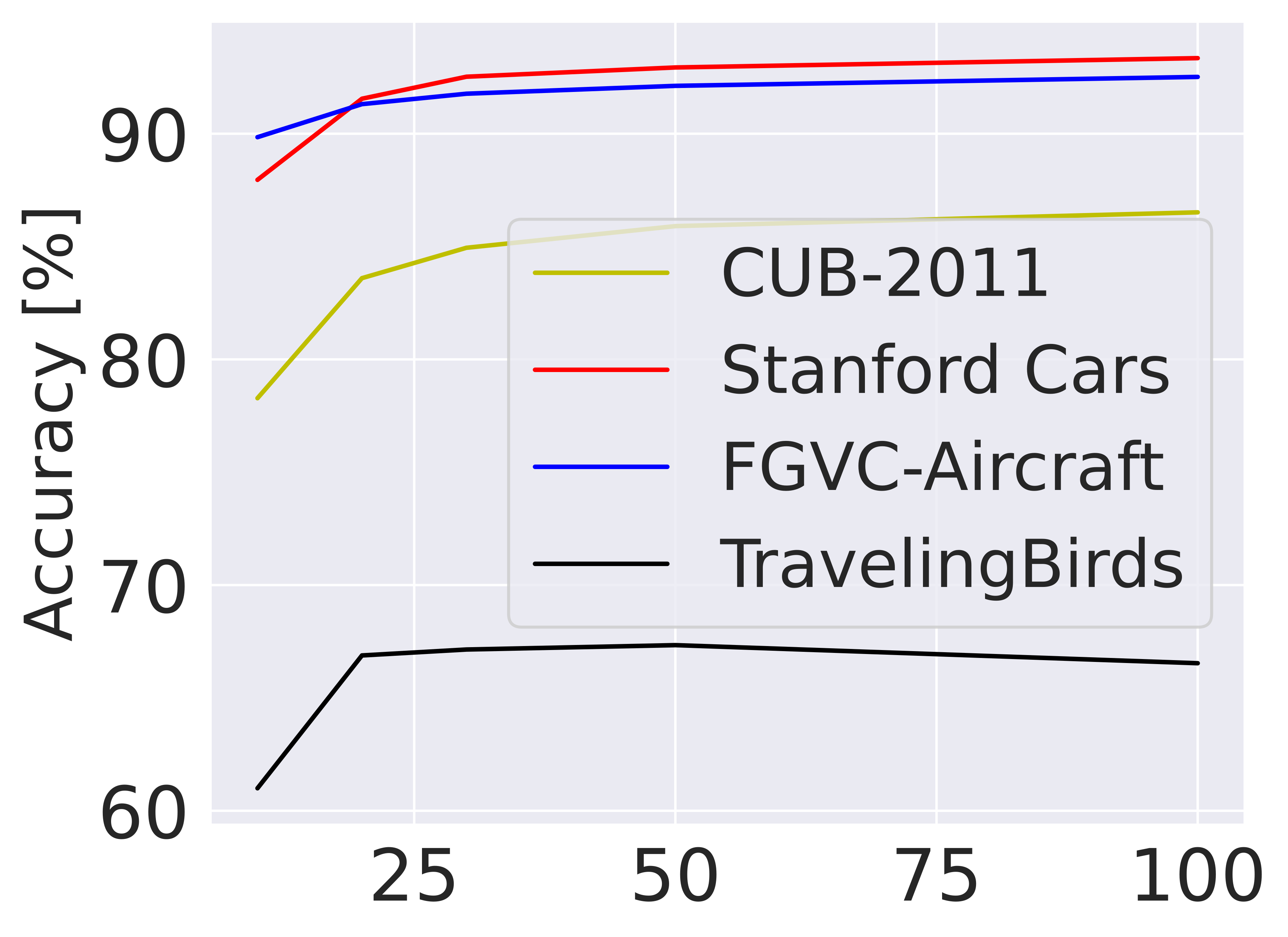}
         \caption{Impact of \gls{nReducedFeatures}}
         \label{fig:NFeaturesImpact}
     \end{subfigure}
     \hfill
     \begin{subfigure}[t]{.49\linewidth}
         \centering
         
       \includegraphics[width=\textwidth]{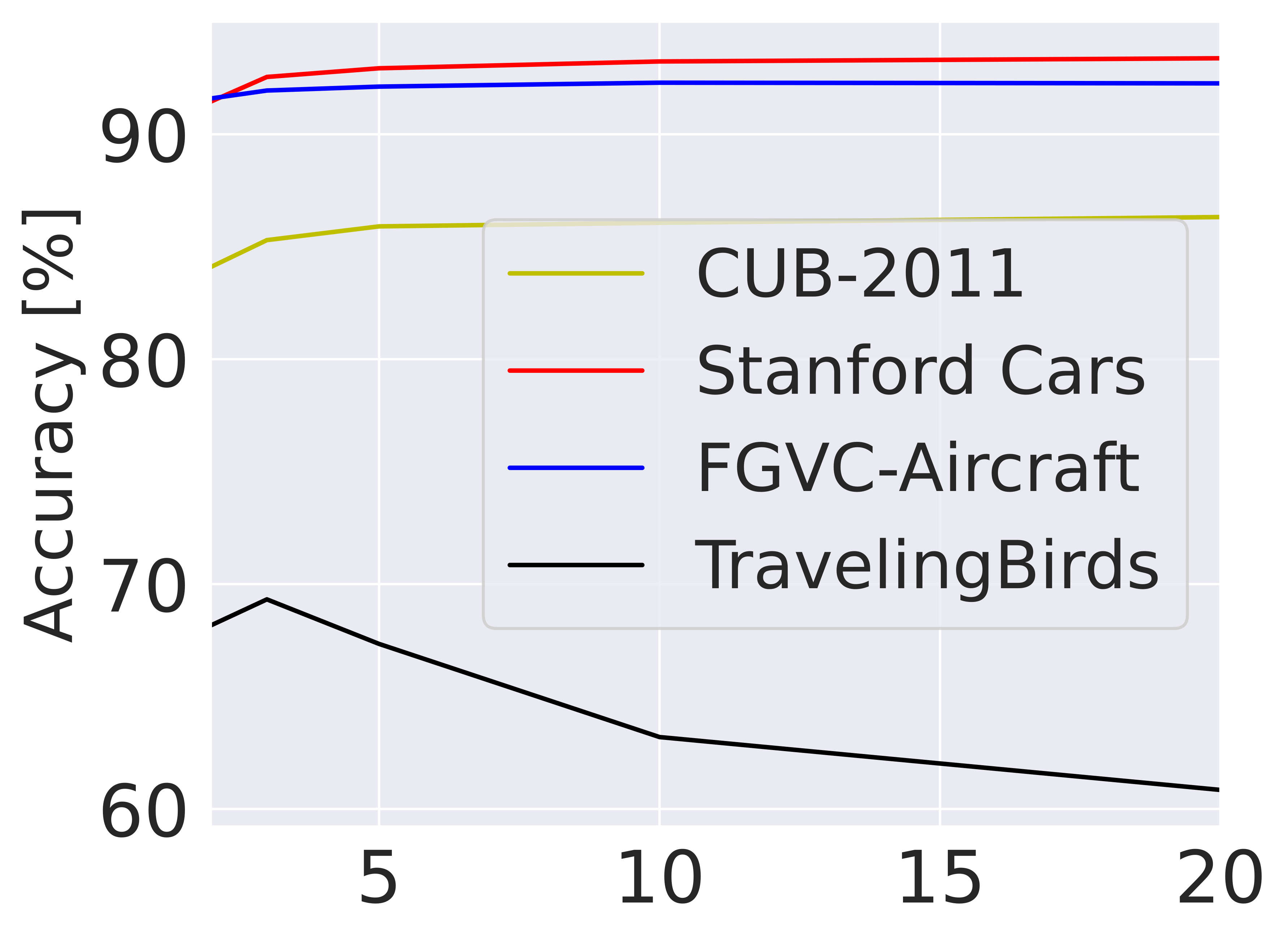}
         \caption{Impact of \gls{nperClass}}
         \label{fig:Impactnwc}
     \end{subfigure}
       \hfill
        \caption{Relationship between Accuracy and interpretability-related parameters  for \gls{NewlayerName} with \gls{resNet}. }
        \label{fig:ContinousInt}
\end{figure}
This section analyzes the impact of changing \gls{nReducedFeatures} and \gls{nperClass} for our~\gls{NewlayerName}. The result is visualized in Figures~\ref{fig:NFeaturesImpact} and~\ref{fig:Impactnwc}:
More features~\gls{nReducedFeatures} and to a lesser degree more weights per class~\gls{nperClass} lead to increased accuracy. However, for \travelingheader{} there is no tradeoff. In fact, more interpretability, especially through sparsity, leads to a higher accuracy because the model does not have the capacity to learn all the spurious correlations but has to focus on the more general features.
Further ablations regarding the convergence of~\gls{qWeightMatrix}, computational cost, type of binary feature and impact
of 
quantization levels are shown in the \supplm. 

\section{Alignment Without Annotations}
\label{sec:FeatureAlignment}
In this section, we propose an alternative way of aligning the learned features of the~\gls{NewlayerName} with human concepts, when no additional annotations are available.
Note that it is easier for \gls{NewlayerName}, as its features show \grounding{}, thus are more alignable with any human concept.
CLIP~\cite{radford2021learning} is utilized as it can compute the similarity between a given text prompt and an image due to its contrastive training. 
The method is first described and then validated by demonstrating high agreement between alignments computed from additional labels provided in \cubheader{} and via our method. 
Finally, we show that it even indicates certainty.
\subsection{CLIP}
In this work, we utilize CLIP~\cite{radford2021learning}, which is a multimodal neural network designed to map images and captions into a shared feature space. The model is trained using a contrastive loss function, which encourages high cosine similarity between the embedding of an image \clipembedding{\gls{ImageSample}}$\in\mathbb{R}^{ \nclipfeatures}$ and its corresponding caption \clipembedding{T}$\in\mathbb{R}^{\nclipfeatures}$, while minimizing the cosine similarity $\clipembedding{\gls{ImageSample}}\clipembedding{T}$ between that image and other captions in the training set. This approach enables the model to learn joint representations that capture meaningful relationships between images and their associated textual descriptions, facilitating downstream tasks such as image captioning and retrieval.
We used the ViT-B32~\cite{dosovitskiy2021an} model of CLIP in our experiments.
\subsection{Feature Alignment}
\begin{figure}
\centering
        \includegraphics[width=\linewidth]{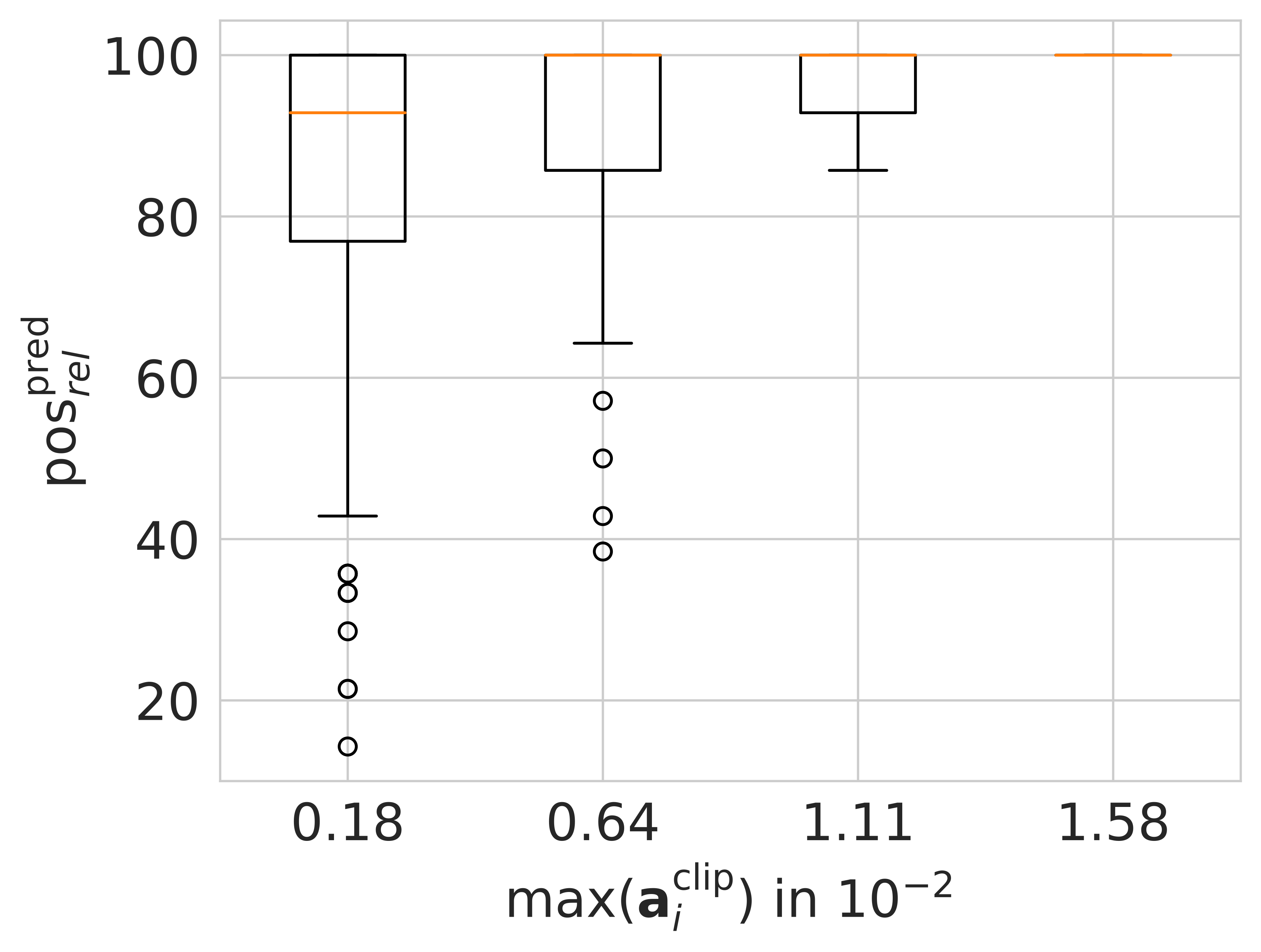}\caption{Relationship between the alignment value $\operatorname{max}(\clipvector_{i})$ and $\posmetric{pred}_{rel}$ for the proposed~\gls{NewlayerName} for all features over $5$ seeds: With increasing $\operatorname{max}(\clipvector_{i})$, the predicted alignment for the feature $i$ is more similar to the ground truth.  }
        \label{fig:relationClip}
\end{figure}

For alignment without annotations, we use the features calculated on the training set~\gls{trainFeatures}$\in\mathbb{R}^{\gls{nTrainImages} \times \gls{nReducedFeatures}}$ and a set of $n_{p}$ possible relevant concepts $a$ for the model such as the attributes contained in~\cubheader. 
The proposed method can identify which of the prompts might be related to the activation of a specific feature. 
First, the entire training set is embedded using CLIP to obtain
$\clipembedding{\mathbf{D}}\in\mathbb{R}^{\gls{nTrainImages}\times \nclipfeatures}$. 
Afterwards, the possibly relevant attributes are converted  into prompts, \eg{}, “This is a photo of a red wing of a bird” for the attribute “red wing” of type "wing color". The conversion is described in detail in
the \supplm{}. 
These prompts are then encoded to $\clipembedding{\mathbf{P}}\in\mathbb{R}^{n_{p}\times \nclipfeatures}$.
The multiplication
\begin{align}
    \mathbf{S} = \clipembedding{\mathbf{D}}\clipembedding{\mathbf{P}}^T
\end{align}
leads to the similarity matrix $\mathbf{S}\in\mathbb{R}^{\gls{nTrainImages}\times n_{p}}$ between images and prompts. In order to relate the image similarities to the learned features, we use the feature values to weight the image similarity according to the activation and obtain one weighting factor $v_{i,j}$
\begin{align}
    v_{ij} =  \frac{\gls{trainFeatures}_{i,j} - mean(\gls{trainFeatures}_{:,j})}{std(\gls{trainFeatures}_{:,j})}
\end{align}
for image $i$ and feature $j$. With this normalized weighting $\mathbf{V}$ the alignment matrix $\mathbf{\clipmatrix}\in\mathbb{R}^{\gls{nReducedFeatures}\times n_{p}}$ is obtained:
\begin{align}
    \mathbf{\clipmatrix} = \mathbf{V}^T\mathbf{S}\quad.
\end{align}
The matrix can be used similarly to $\mathbf{\gtmatrix}$ as it measures how much the perceived similarity to a specific prompt of CLIP varies along the feature dimension.

Note that the proposed method is generally independent of the used CLIP-like model, and could therefore profit of models that are able to caption more fine-grained subregions.
\subsection{Validation}
\begin{table}
\resizebox{\linewidth}{!}{
\begin{tabular}{l|c|c|c} 
  \toprule Alignment Method   & $\posmetric{full} $\arrowDown &$\posmetric{pred}$ \arrowDown  & $\posmetric{pred}_{rel}$ \arrowUp \\
  \midrule  
   Random  & 156.5 & 7.3 & 50.0\% \\
        Static Sorting with minimal \posmetric{full}& 122.9 & 8.0 & 41.7\% \\
\hdashline
  Proposed Method&   \textbf{43.1} & \textbf{2.7} & \textbf{86.2}\% \\
  \bottomrule
\end{tabular}
}
\caption{Average validation metrics when aligning features of \gls{NewlayerName} compared to baselines: Our proposed alignment method handily beats the optimal static baseline based on the ground 
truth.
Rather than just identifying relevant concepts, our proposed method matches it to the correct feature.}
\label{tab:AvgPosition}
\end{table}
For validation, we test how well our method can identify the most related concept from a given set of concepts. 
As ground truth, the alignment matrix $\mathbf{\gtmatrix}$ computed 
with Equation~\ref{eq:Agt}
is used. 
To compare whether our method can \textit{order} the attributes in the same way, we use all attributes contained  in~\cubheader{} as set of possible attributes and compute $\mathbf{\clipmatrix}$.
For each feature $i$ $\posmetric{full}_i$ then measures the position at which the most aligned attribute $a$ in $\clipvector_i$ is ranked in $\gtvecvtor_i$:
\begin{align}
\posmetric{full}_i = \{j : \operatorname{argmax}(\clipvector_i) = \operatorname{argsort}(\gtvecvtor_i)_j\}\quad.
\end{align}
Here, $\operatorname{argmax}$ returns the index with the highest value and $\operatorname{argsort}(\gtvecvtor_i)$ returns the indices that would sort $\gtvecvtor_i$ in descending order.
We
additionally calculate the above metric for the predicted type by limiting \gtvecvtor{} and \clipvector{} to its attributes.
It focuses on the ability to identify which expression of a given type is more related to a feature, while  \posmetric{full} measures if the method can identify type and expression, which can be more affected by biases of CLIP.
To be less dependent on the number of attributes of the predicted type $n_{type}$, we also report the relative position
\begin{equation}
    \posmetric{pred}_{rel, i} = 1 - \frac{\posmetric{pred}_{i} -1}{n_{type} -1} = \frac{n_{type}-\posmetric{pred}_{i}}{n_{type} -1},
\end{equation}
which ranges from the worst case at 0\% 
to
optimality
at 100\%.
The results 
in 
Table~\ref{tab:AvgPosition} 
validate the proposed method and show the superiority to any static baseline.
\subsubsection{Optimal Static Baseline}
To compute the optimal static sorting of attributes for comparison with our alignment method, we calculated $pos_{i,j} = \operatorname{argsort}(\gtvecvtor_i)_j$ for all features $i$ and attributes $j$ and sorted the attributes, \st{} the average position
\begin{equation}
    avg_j = \frac{1}{\gls{nReducedFeatures}} \sum_{i=0}^{\gls{nReducedFeatures}}pos_{i,j}
\end{equation}
of the attribute ascends.
This returns the optimal static list which minimizes \posmetric{full}.
\subsection{Analysis of Alignment Value}
Figure~\ref{fig:relationClip} shows the relationship between the alignment value $\operatorname{max}(\clipvector_{i})$ and 
$\posmetric{pred}_{rel}$
for the proposed~\gls{NewlayerName} for all features over $5$ seeds. The box plots are created by assigning every feature to one of the $4$ evenly sized bins according to their alignment value $\operatorname{max}(\clipvector_{i})$.
The other metrics show similar trends.
The figure demonstrates that our method conveys certainty, as $\posmetric{pred}_{rel}$ increases with $\operatorname{max}(\clipvector_{i})$.
Since the features do not need to be aligned with any of the given attributes, Figure~\ref{fig:relationClip} suggests that the features with higher uncertainty 
do not directly correspond to any of the given attributes, but rather concepts not annotated in \cubheader{}.
Therefore, the proposed method with no required additional annotation is more accurate for features aligned with the given attributes than the reported average in Table~\ref{tab:AvgPosition} suggests, with the median absolute value of \posmetric{pred} for the top $3$ bins being the first position, shown in Figure~\ref{fig:relationClipAbs}.

\section{Future Work}
\gls{NewlayerName} follows the popular SENN~\cite{alvarez2018towards} framework and delivers concise, contrastive and general, therefore human-friendly~\cite{miller2019explanation}, explanations, as shown in Figure~\ref{fig:Cover}. 
However, human-subject experiments are required to actually measure the interpretability for humans~\cite{doshi2017towards}. 
As \gls{NewlayerName} is generally applicable to classification tasks, it can also bring more interpretability to other tasks like semantic segmentation, where interpretability is discussed~\cite{kaiser2023compensation}.
\section{Conclusion}
We propose the
\textit{Quantized-Self-Explaining Neural Network}~\gls{NewlayerName} 
with iteratively optimized
ternary
feature-class 
assignments.
The experiments support the classification
as SENN by evaluating how the concepts 
learned without additional supervision 
satisfy the 
three 
desiderata \fidelity, \diversity{} and \grounding{}.
In isolation, quantization primarily benefits \fidelity{}, dependence~\dependence{} and \diversity{}, while 
just
iterating increases feature alignment $r$ and robustness to spurious correlations.
\gls{NewlayerName} combines and amplifies the individual strengths,
surpassing competitors in various facets.
Finally, we propose a method for aligning the learned concepts with human ones with no need for additional supervision using CLIP, enabling~\gls{NewlayerName} to be applied in various scenarios with need for interpretability.
Future work might even incorporate it
to guide \gls{NewlayerName} towards more grounded concepts.

 \section{Acknowledgements}
This work was supported by the Federal Ministry of Education and Research (BMBF), Germany under the AI service center KISSKI (grant no. 01IS22093C) and the Deutsche Forschungsgemeinschaft (DFG) under Germany’s Excellence Strategy within the Cluster of Excellence PhoenixD (EXC 2122). 
This work was partially supported by Intel Corporation and by the German Federal Ministry
of the Environment, Nature Conservation, Nuclear Safety
and Consumer Protection (GreenAutoML4FAS project no.
67KI32007A). 

\bibliography{aaai24}

\appendix
\clearpage
\twocolumn[
  \begin{@twocolumnfalse}
    \begin{center}
    \Large\textbf{Supplementary Material}
    \bigskip
\end{center}
  \end{@twocolumnfalse}
]

\section{Introduction}
The supplementary material contains more detailed results and examples, implementation details, the visualization techniques and ablation studies regarding the convergence during iteration, relation  between the alignment value $\operatorname{max}(\clipvector_{i})$ and  \posmetric{pred}, number of quantization levels and the relation between \loc{5} and correlation. 
Additionally, the computational cost of \gls{NewlayerName} is discussed.

\section{Implementation Details}
\label{app:ImpDetails}
We utilize PyTorch~\citep{Pytorch} for implementing our methods and use \gls{imgnetheader} pretrained models, as backbone feature extractor. We use \glm{} and the \textit{robustness}~\citep{robustness} framework. Generally, the implementation details are similar to the \glsname{layerName}~\citep{norrenbrocktake}.
To prepare the images for training, we resize them to $448\times448$ ($299\times299$ for \incv) and applied  normalization, random horizontal flip, random crop with a padding of $4$, jitter and \textit{TrivialAugment}~~\cite{Muller_2021_ICCV}. \textit{TrivialAugment}~ was omitted for ImageNet.
We left out the crop for \fgvcheader, as the aircraft usually span the entire image.
For comparison with prototype-based models, we used the usual prototype data, i.\,e. for \gls{cubheader} the images are cropped based on a ground-truth segmentation and all images are resized to $224\times224$. 
Additionally, we set the strides of the final 2 blocks to 1, following~\gls{PIP-Net}.

The dense models are fine-tuned on the dataset using stochastic gradient descent with a batch size of $16$ for $150$ epochs, with a starting learning rate of $5\cdot10^{-3}$ for the pretrained layer and $0.01$ for the final linear layer. The learning rates are multiplied by $0.4$ every $30$ epochs. We use momentum of $0.9$, $\ell_2$-regularization of $5\cdot10^{-4}$, and a dropout rate of $0.2$ on the features to reduce dependencies. \gls{cLW} is set to $0.196$ for \resnet{}, $0.098$ for \densenet{}, and $0.049$ for \incv.
To perform feature selection, we set $\gls{elaW}$ to 0.8 and decrease the regularization strength, \gls{elaWeight}, by 90. 
We utilize \glm{} to calculate the regularization path with $\gls{elaW}=0.99$ and all other parameters set to default, including a lookbehind of $T=5$.
As discussed in the  section on computational cost, we use an accelerated setting for~\gls{NewlayerName} on \gls{imgnetheader}. with higher tolerance and shorter path length.
Then we apply the quantization to the least regularized matrix in the regularization path to obtain a matrix with $\gls{nperClass}=5$.
After that, the model is trained with the final layer fixed to the sparse quantized solution for $10$ epochs. The learning rate for the sparse fine-tuning is set to $100$ times the final learning rate of the dense training, but is reduced by $10\%$ at each iteration.
The cycle of calculating the sparse matrix and fine-tuning is repeated $N=4$ times with the last fine-tuning taking $40$ epochs with a decrease of the learning rate by $60\%$ every $10$ epochs.
For all fine-tuning iterations, dropout is set to $0.1$ and momentum to $0.95$.
For every experiment, all random seeds are set to a random integer between $1$ and $10^7$, drawn using the \textit{Python}~\citep{10.5555/1593511} \textit{random} library.

For evaluating the alignment $r$ we considered all attributes to be present, i.\,e. part of $\attributeset{a+}$, that were annotated as present with \textit{probably} or higher as annotated certainty. 

The box plots in Figure~\ref{fig:relationClip} are created using the default parameters from \textit{matplotlib}~\cite{Hunter:2007}
\subsection{Prompts}
\label{sec:Prompts}
The attributes in \glsentrylong{cubheader} have the form \textit{type: expression}, \eg{}, \textit{eye-color: green}. To convert them into prompts for CLIP~\cite{radford2021learning}, we use the following template: 

"This is a photo of a \textit{expression} \textit{type} of a bird."

Some attributes do not fit the template, because the expression is no adjective, \eg{} needle.
For those, we used the template:

"This is a photo of a \textit{expression}-like \textit{type} of a bird."

Note, that the prompts are very simple and can be further optimized, \eg{}, adapted to the type.

\subsection{Clustering}
\begin{figure}
\begin{subfigure}[t]{.45\linewidth}
     \includegraphics[width=\textwidth]{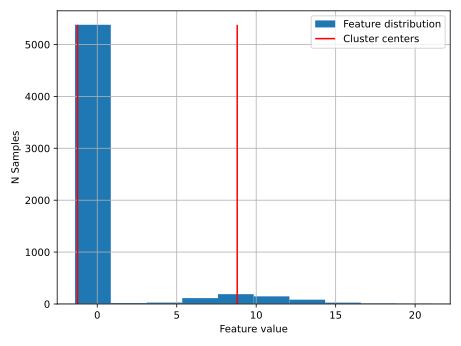}
         \caption{\gls{NewlayerName}}
     \end{subfigure}
\hfill
     \begin{subfigure}[t]{.45\linewidth}
         \centering
         \includegraphics[width=\textwidth]{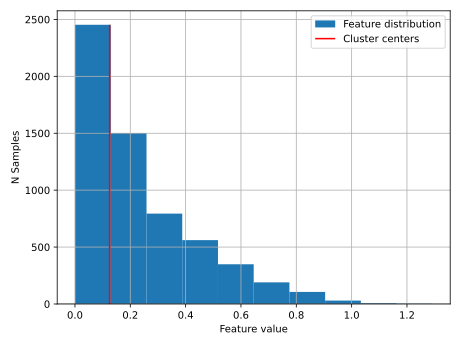}
         \caption{Dense \resnet{}}
     \end{subfigure}

      \hfill
        \caption{Exemplary clustering of one feature for our~\gls{NewlayerName} compared to a dense model on~\gls{cubheader}: \gls{NewlayerName} learns more binary features.}
        \label{fig:ClusterEx}
\end{figure}

This section describes the clustering methodology used to obtain the fraction of more binary features.
For that, we applied mean shift clustering on the distribution of the features over the entire training set.  
Visual examples are given in \suppl{} Figure~\ref{fig:ClusterEx}.
The used bandwidth is the default for \textit{scikit-learn}~\cite{scikit-learn}, which is the 30th percentile of the pairwise distances.
On top of the binary-like percentage of the main paper, \suppl{} Table~\ref{stable:binaryReal} contains the percentage of features that are clustered in 2 groups and where  both clusters have at least as many entries as the average class has training examples.
Interestingly, while \gls{cbm} and \gls{NewlayerName} learn actual binary concepts, the binary-like features of the dense model are class-specific.
\subsection{Competitors}
\label{app:cbmjoint}
To generate \textit{\gls{cbm} - joint} models with \resnet{}, we resize the images to $448\times448$ and use a batch size of $16$. Almost all remaining parameters are identical to the \incv{} experiments of \glsname{cbm}, but we also used \textit{TrivialAugment}~\cite{Muller_2021_ICCV} as it improved performance for our experiments.
For alignment $r$ and the binary-like features metric we first applied the sigmoid function to the features, as they were trained with that activation.
To generate results for prototype-based models on \gls{travelingheader}, we also cropped the images according to the ground truth segmentation and else used the parameters reported for \cubheader{}. 
For \gls{ProtoPool} however, we also used an \imgnetheader{} pretrained model for better comparability and omitted Mix-Up~\cite{thulasidasan2019mixup}, as it likely affects the sensitivity to background changes.

\section{Convergence of \glsentryname{qWeightMatrix}}
\begin{figure}
\begin{center}
  \includegraphics[width=\linewidth]{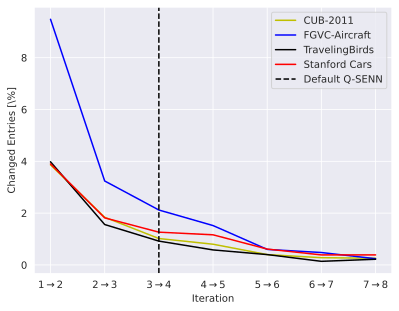}
\end{center}
  \caption{Fraction of changed entries in~\glsentryname{qWeightMatrix} while iterating}
  \label{fig:iterConv}
 \end{figure}
In this section, we investigate the convergence of \glsentryname{qWeightMatrix}.
For that, we trained \glsentryname{NewlayerName} with $N=8$ iterations and measured how many of the assignments changed between cycles.
Specifically, \suppl{} 
Figure~\ref{fig:iterConv} shows the number of new assignments between features and classes as fraction of total assignments.
The continuous decrease with iterations is apparent, validating our hypothesis of a converging \glsentryname{qWeightMatrix}. 
We chose $N=4$ to save resources, as most of the adaptation happens until that iteration.

\section{Number of Quantization Levels}
   \begin{figure*}
          \begin{subfigure}[t]{.45\textwidth}
       \centering
     \includegraphics[width=\linewidth]{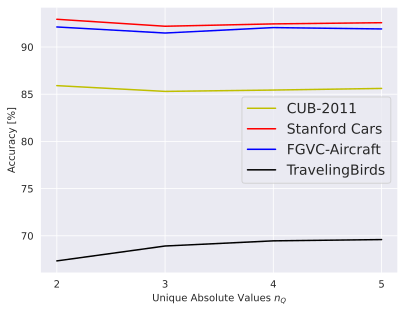}
  \caption{On Accuracy}
         \label{fig:ImpactUniquesAcc}  
     \end{subfigure} \hfill
      \begin{subfigure}[t]{.45\textwidth}
       \centering
     \includegraphics[width=\linewidth]{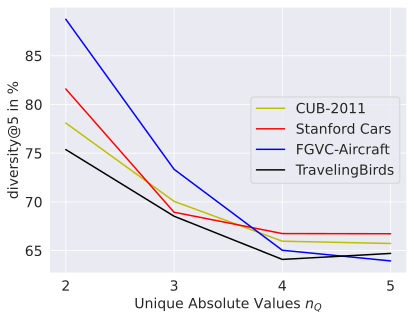}
  \caption{On \loc{5}}
         \label{fig:ImpactUniquesDiv}  
     \end{subfigure}    \caption{Impact of  $n_Q$ for \gls{NewlayerName} with \resnet{}}
         \label{fig:ImpactUniques}
     \end{figure*}
In this section, the impact of the number of unique values $n_{Q}$ in the quantized matrix~\glsname{qWeightMatrix} is analyzed.
 
\begin{figure}
\begin{center}
  \includegraphics[width=\linewidth]{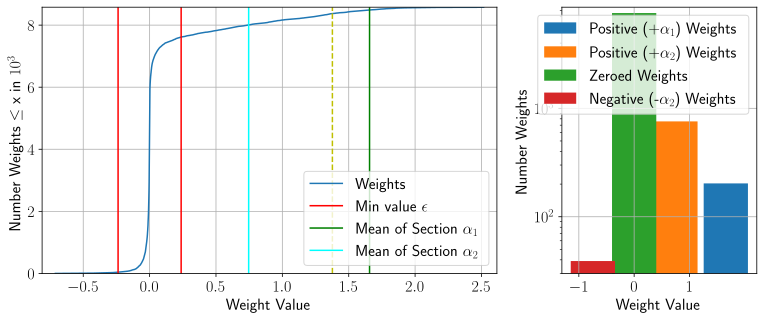}
\end{center}
  \caption{Exemplary result (right) of quantization on cumulative distribution (left) of nonzero weights in  \sparsemat{} for \cubheader($\gls{nWeights}=1000)$ for $n_Q=3$}
  \label{fig:nq3}
 \end{figure}

\begin{figure}
\begin{center}
  \includegraphics[width=\linewidth]{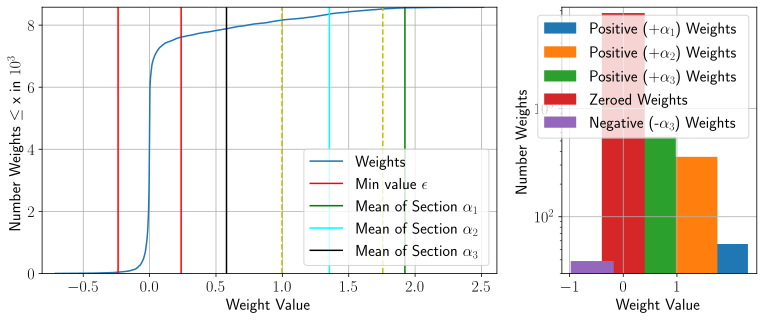}
\end{center}
  \caption{Exemplary result (right) of quantization on cumulative distribution (left) of nonzero weights in  \sparsemat{} for \cubheader($\gls{nWeights}=1000)$ for $n_Q=4$}
  \label{fig:nq4}
 \end{figure}

For that, we split the range of values above~$\epsilon$ into $n_{Q}-1$ evenly sized bins and set each value in each bin to the respective average, visualized in \suppl{} Figures~\ref{fig:nq3} and~\ref{fig:nq4}  for $n_{Q} = 3,4$.
\suppl{} Figure~\ref{fig:ImpactUniquesAcc} shows that increasing the number of unique values does not generally increase accuracy. A slight improvement is only obtained for \travelingheader{}. 
This demonstrates how increasing the interpretability by calculating assignments does not negatively impact the accuracy. 
Notably, the \loc{5} decreases strongly with $n_{Q}$, shown in \suppl{} Figure~\ref{fig:ImpactUniquesDiv}. 
It is already similar to the~\gls{layerName} with $n_{Q}=4$.
This is an expected behavior, as the quantization initially increases~\loc{5} and increasing $n_Q$ effectively reduces quantization.


\section{Visualization Techniques}
\label{sec:VizTec}
For visualization, we used Grad-CAM~\citep{selvaraju2017grad} for Figure 4 in the main paper. 
For \suppl{} Figures~\ref{app:fig:vizExamplesCardinal} to~\ref{app:fig:vizExamplesE-170} we overlayed the feature maps with color map \textit{jet} over the images.
For the global explanation of~\gls{NewlayerName} in Figure~1 in the main paper, we converted the images to grayscale, so that feature maps are more visible. After that, we overlayed the images with the respective feature maps. We also applied gamma correction to the feature maps so that they match the visual impression the \textit{jet} color map (\suppl{} Figure~\ref{app:fig:vizExamplesWestern_Wood_Pewee}) gives.

For the local explanations of \gls{NewlayerName} in Figures~1 and~4 in the main paper, we only overlaid pixels where any of the features was above $70\%$ of its maximum activation for the image. To make the image clearer, we only showed the maximum feature per pixel.

\section{Detailed Results}

\begin{figure}
\begin{center}
  \includegraphics[width=\linewidth]{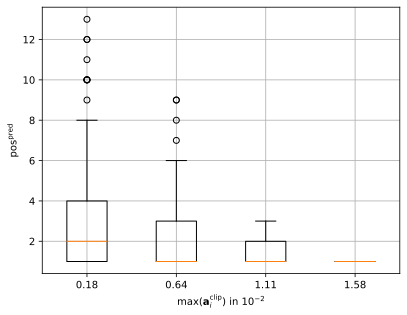}
\end{center}
  \caption{Relationship between the alignment value $\operatorname{max}(\clipvector_{i})$ and \posmetric{pred} for the proposed~\gls{NewlayerName} for all features over $5$ seeds: With increasing $\operatorname{max}(\clipvector_{i})$, the predicted alignment for the feature $i$ is more similar to the ground truth. }
  \label{fig:relationClipAbs}
  \vspace{.1cm}
 \end{figure}

\begin{table*}
\resizebox{\linewidth}{!}{
\centering
\begin{tabular}{c|ccc|ccc|ccc|ccc}
\toprule
&\multicolumn{3}{c|}{CUB-2011}&\multicolumn{3}{c|}{FGVC-Aircraft}&\multicolumn{3}{c|}{Stanford Cars}&\multicolumn{3}{c}{TravelingBirds}\\
Method & \sinception & \sdensenet & \sresnet  & \sinception & \sdensenet & \sresnet & \sinception & \sdensenet & \sresnet & \sinception & \sdensenet & \sresnet \\
\toprule

\densetable{}& {83.5}$\pm$0.3 & {87.2}$\pm$0.2 & {86.7}$\pm$0.2 & {91.4}$\pm$0.2 & {92.5}$\pm$0.3 & {92.0}$\pm$0.2 & {92.6}$\pm$0.2 & {93.6}$\pm$0.1 & {93.5}$\pm$0.1 & {47.9}$\pm$0.5 & {45.1}$\pm$0.8 & {39.4}$\pm$0.4 \\

\midrule

\glmtable{} & 76.9$\pm$0.3 & 69.8$\pm$1.3 & 77.7$\pm$0.4 & 89.1$\pm$0.3 & 86.2$\pm$0.5 & 89.9$\pm$0.6 & 88.9$\pm$0.2 & 81.9$\pm$0.4 & 89.1$\pm$0.5 & 45.7$\pm$0.6 & 42.7$\pm$1.2 & 36.0$\pm$0.8 \\

\gls{cbm}-joint & 80.1$\pm$0.6 & - & 82.2$\pm$0.2  & N/A & N/A&N/A   & N/A &  N/A& N/A& 51.8$\pm$1.8  & - & 36.9$\pm$0.3 \\
SLDD-Model & {80.8}$\pm$0.3 & {84.4}$\pm$0.3 & {85.7}$\pm$0.1 & {90.6}$\pm$0.1 & {92.1}$\pm$0.2 & {92.0}$\pm$0.2 & {91.3}$\pm$0.2 & {92.2}$\pm$0.2 & \textbf{92.9}$\pm$0.1 & \textbf{51.9}$\pm$0.2 & {56.1}$\pm$0.5 & {64.1}$\pm$0.4 \\

\hdashline
Q-SENN (Ours) & \textbf{81.7}$\pm$0.1 & \textbf{85.4}$\pm$0.2 & \textbf{85.9}$\pm$0.4 & \textbf{90.8}$\pm$0.2 & \textbf{92.5}$\pm$0.2 & \textbf{92.1}$\pm$0.2 & \textbf{91.8}$\pm$0.2 & \textbf{92.8}$\pm$0.1 & \textbf{92.9}$\pm$0.1 & 50.5$\pm$0.5 & \textbf{59.6}$\pm$0.8 & \textbf{67.3}$\pm$0.5 \\
\bottomrule
\end{tabular}
}
\caption{\fidelity{} measured as accuracy\tablefinisher{backbone}. 
Best result among more interpretable models are in bold. 
}
\label{stable:Fidelity}
\end{table*}
\begin{table*}
\resizebox{\linewidth}{!}{
\centering
\begin{tabular}{c|ccc|ccc|ccc|ccc}
\toprule
&\multicolumn{3}{c|}{CUB-2011}&\multicolumn{3}{c|}{FGVC-Aircraft}&\multicolumn{3}{c|}{Stanford Cars}&\multicolumn{3}{c}{TravelingBirds}\\
Method & \sinception & \sdensenet & \sresnet  & \sinception & \sdensenet & \sresnet & \sinception & \sdensenet & \sresnet & \sinception & \sdensenet & \sresnet \\
\toprule
\densetable{} & {39.1}$\pm$0.3 & {65.8}$\pm$0.3 & {51.6}$\pm$0.2 & {35.8}$\pm$0.3 & {66.8}$\pm$0.4 & {46.7}$\pm$0.5 & {35.8}$\pm$0.4 & {64.2}$\pm$0.3 & {47.6}$\pm$0.2 & {37.5}$\pm$0.2 & {64.9}$\pm$0.2 & {51.5}$\pm$0.2 \\

\midrule
\glmtable{} & {39.4}$\pm$0.8 & {70.6}$\pm$0.2 & {46.7}$\pm$0.3 & {37.4}$\pm$0.4 & {71.3}$\pm$0.9 & {43.1}$\pm$0.5 & {36.0}$\pm$0.3 & {73.7}$\pm$0.4 & {43.7}$\pm$0.4 & {39.1}$\pm$0.2 & {70.9}$\pm$0.3 & {47.7}$\pm$0.3 \\

\gls{cbm}-joint & - & - & 71.0$\pm$0.8  & N/A & N/A&N/A   & N/A &  N/A& N/A& -  & - & 73.7$\pm$0.9 \\
SLDD-Model  & {43.7}$\pm$0.3 & \textbf{80.3}$\pm$0.7 & {67.2}$\pm$1.6 & {51.0}$\pm$1.3 & \textbf{85.5}$\pm$1.0 & {75.6}$\pm$1.8 & {42.5}$\pm$0.8 & \textbf{82.6}$\pm$1.4 & {68.4}$\pm$1.4 & {41.4}$\pm$0.8 & {78.6}$\pm$0.6 & {65.8}$\pm$1.1 \\

\hdashline

Q-SENN (Ours) & \textbf{52.9}$\pm$0.8 & \textbf{80.3}$\pm$1.0 & \textbf{78.1}$\pm$1.5 & \textbf{57.9}$\pm$1.6 & {82.3}$\pm$1.5 & \textbf{88.7}$\pm$0.8 & \textbf{52.8}$\pm$1.1 & {81.0}$\pm$1.3 & \textbf{81.6}$\pm$1.0 & \textbf{50.2}$\pm$0.8 & \textbf{79.2}$\pm$1.4 & \textbf{75.4}$\pm$0.9 \\


\bottomrule

\end{tabular}
}
\caption{\diversity{} measured as \loc{5}\tablefinisher{backbone}. 
Best result among more interpretable models are in bold. 
}
\label{stable:Diversity}
\end{table*}

\begin{table*}
\resizebox{\linewidth}{!}{
\centering
\begin{tabular}{c|ccc|ccc|ccc|ccc}
\toprule
&\multicolumn{3}{c|}{CUB-2011}&\multicolumn{3}{c|}{FGVC-Aircraft}&\multicolumn{3}{c|}{Stanford Cars}&\multicolumn{3}{c}{TravelingBirds}\\
Method & \sinception & \sdensenet & \sresnet  & \sinception & \sdensenet & \sresnet & \sinception & \sdensenet & \sresnet & \sinception & \sdensenet & \sresnet \\
\toprule

\densetable{ }  & {24.0}$\pm$1.2 & {16.6}$\pm$0.6 & {14.6}$\pm$1.2 & {30.8}$\pm$2.7 & {16.6}$\pm$0.6 & {19.1}$\pm$1.9 & {14.6}$\pm$1.0 & {11.7}$\pm$0.9 & {5.0}$\pm$0.6 & {22.2}$\pm$1.1 & {15.5}$\pm$1.2 & {15.0}$\pm$0.9 \\

\midrule

\glmtable{}  & {21.4}$\pm$0.7 & {31.2}$\pm$1.0 & {12.6}$\pm$2.0 & {33.5}$\pm$2.2 & \textbf{31.5}$\pm$0.6 & {19.6}$\pm$2.3 & {19.2}$\pm$0.9 & {22.4}$\pm$1.9 & {4.5}$\pm$1.0 & {21.9}$\pm$1.5 & {30.5}$\pm$2.6 & {13.3}$\pm$1.0 \\

\gls{cbm}-joint & - & - & 67.0$\pm$1.5  & N/A & N/A&N/A   & N/A &  N/A& N/A& -  & - & 45.5$\pm$3.9 \\
SLDD-Model & {40.8}$\pm$6.0 & {7.6}$\pm$2.0 & {78.0}$\pm$1.3 & {29.6}$\pm$4.5 & {29.7}$\pm$5.9 & \textbf{74.0}$\pm$6.1 & {43.6}$\pm$6.4 & {4.4}$\pm$2.0 & {74.4}$\pm$5.0 & {36.4}$\pm$5.9 & {6.0}$\pm$3.1 & {55.2}$\pm$4.5 \\
\hdashline
Q-SENN (Ours) & \textbf{53.2}$\pm$5.2 & \textbf{51.6}$\pm$3.4 & \textbf{80.4}$\pm$8.7 & \textbf{37.2}$\pm$6.9 & {31.4}$\pm$4.2 & {71.2}$\pm$8.1 & \textbf{56.4}$\pm$2.9 & \textbf{56.0}$\pm$2.8 & \textbf{90.0}$\pm$4.4 & \textbf{50.8}$\pm$6.8 & \textbf{44.4}$\pm$10.4 & \textbf{84.0}$\pm$2.5 \\
\bottomrule

\end{tabular}
}
\caption{Binary Features in percent\tablefinisher{backbone}. 
Best result among more interpretable models are in bold. 
}
\label{stable:binary}
\end{table*}
\begin{table*}
\resizebox{\linewidth}{!}{
\centering
\begin{tabular}{c|ccc|ccc|ccc|ccc}
\toprule
&\multicolumn{3}{c|}{CUB-2011}&\multicolumn{3}{c|}{FGVC-Aircraft}&\multicolumn{3}{c|}{Stanford Cars}&\multicolumn{3}{c}{TravelingBirds}\\
Method & \sinception & \sdensenet & \sresnet  & \sinception & \sdensenet & \sresnet & \sinception & \sdensenet & \sresnet & \sinception & \sdensenet & \sresnet \\
\toprule

\densetable{} & {0.7}$\pm$0.2 & {0.1}$\pm$0.0 & {0.0}$\pm$0.0 & {2.1}$\pm$0.3 & {0.1}$\pm$0.1 & {0.2}$\pm$0.1 & {1.5}$\pm$0.2 & {0.1}$\pm$0.1 & {0.1}$\pm$0.1 & {0.8}$\pm$0.1 & {0.1}$\pm$0.1 & {0.1}$\pm$0.1 \\

\midrule

\glmtable{} & {3.0}$\pm$0.5 & {0.2}$\pm$0.1 & {0.1}$\pm$0.1 & {6.3}$\pm$0.8 & {0.3}$\pm$0.1 & {0.8}$\pm$0.5 & {5.4}$\pm$0.7 & {0.3}$\pm$0.2 & {0.2}$\pm$0.1 & {2.7}$\pm$0.5 & {0.2}$\pm$0.1 & {0.3}$\pm$0.2 \\

\gls{cbm}-joint & - & - & 66.7$\pm$1.8  & N/A & N/A&N/A   & N/A &  N/A& N/A& -  & - & 44.0$\pm$3.4 \\
SLDD-Model & {39.2}$\pm$5.9 & {2.4}$\pm$0.8 & {77.6}$\pm$1.5 & {28.8}$\pm$4.1 & {24.9}$\pm$2.9 & \textbf{73.2}$\pm$5.9 & {43.2}$\pm$5.7 & {4.0}$\pm$2.0 & {74.0}$\pm$5.4 & {31.6}$\pm$4.6 & {2.0}$\pm$1.0 & {52.4}$\pm$4.1 \\

\hdashline
Q-SENN (Ours)  & \textbf{53.2}$\pm$5.2 & \textbf{44.4}$\pm$2.9 & \textbf{80.4}$\pm$8.7 & \textbf{37.2}$\pm$6.9 & \textbf{31.4}$\pm$4.2 & {71.2}$\pm$8.1 & \textbf{56.0}$\pm$3.6 & \textbf{53.2}$\pm$1.6 & \textbf{90.0}$\pm$4.4 & \textbf{50.8}$\pm$6.8 & \textbf{38.8}$\pm$7.8 & \textbf{84.0}$\pm$2.5 \\

\bottomrule

\end{tabular}
}
\caption{Binary Features in percent\tablefinisher{backbone} where every cluster has to have more points than training samples for one class. 
Best result among more interpretable models are in bold. 
}
\label{stable:binaryReal}
\end{table*}

\begin{table*}
\begin{tabular}{l|c|c|c} 
  \toprule Alignment Method   & $\posmetric{full} $\arrowDown &$\posmetric{pred}$ \arrowDown  & $\posmetric{pred}_{rel}$ \arrowUp \\
  \midrule

   Random  & 156.5 & 7.3$\pm$0.1 & 50.0\% \\

        Static Sorting with minimal \posmetric{full}& 122.9$\pm$4.0 & 8.0$\pm$2.3 & 41.7\%$\pm$5.6\% \\

\hdashline
  Proposed Method&   \textbf{43.1}$\pm$5.1 & \textbf{2.7}$\pm$0.3 & \textbf{86.2}\%$\pm$1.9\% \\
  \bottomrule
\end{tabular}
\caption{Average validation metrics when aligning features of \gls{NewlayerName} compared to baselines: Our proposed alignment method handily beats the optimal static baseline based on the ground 
truth.
Rather than just identifying relevant concepts, our proposed method matches it to the correct feature.}
\label{stab:AvgPosition}
\end{table*}
\begin{table*}[htbp]
  \centering
  \begin{tabular}{l|ccc|ccc|ccc}
    \toprule
    Method &  \multicolumn{3}{c|}{Accuracy \arrowUp} &
    \multicolumn{3}{c|}{Total Feat.\arrowDown} & \multicolumn{3}{c}{Feat./ Class\arrowDown} \\
  & C& S & T & C& S & T & C& S &T
    \\
    
    \midrule
    PIP-Net & 82.0$\pm0.3$ & 86.5$\pm0.3$ & 70.1$\pm0.9$  & 731$\pm19$ &669$\pm13$& 825$\pm37$  & 12 &11 & 5.6$\pm0.2$ \\ 
    ProtoPool & 85.5$\pm0.1$ &88.9$\pm0.1$& 42.9$\pm$1.7& 202 & 195 &202 &10 & 10 & 10 \\
     \midrule
     Q-SENN & 84.7$\pm0.3$ & 91.5$\pm0.3$ &76.7$\pm0.5$& \textbf{50} & \textbf{50} &\textbf{50}& \textbf{5}& \textbf{5} &\textbf{5} \\
    $\gls{nReducedFeatures}>50$ & \textbf{86.4}$\pm0.3$ & \textbf{92.2}$\pm0.2$& \textbf{78.4}$\pm0.5$ & 202 &195& 202& \textbf{5} &\textbf{5}  &\textbf{5} \\
 
    \bottomrule
  \end{tabular}
  \caption{Comparison with state-of-the-art prototype-based methods with \resnet{} as backbone on the same data: With reduced sizes, \gls{NewlayerName} shows increased accuracy.}
  \label{stab:proto-table}
  
\end{table*}

In this section, the standard deviations for the presented results are included. 
All metrics are the average of 5 runs, 3 for the experiments with~\glsentryfirst{ProtoPool}, \glsentryfirst{PIP-Net}, \glsentryfirst{cbm} and on \gls{imgnetheader}, with $\pm$ indicating one standard deviation. 
The full results with the standard deviations are presented in
\suppl{}
Tables~\ref{stable:Fidelity} to~\ref{stab:abl}. 
The reported standard deviations are generally rather small compared to the differences in means, which supports our conclusions. 

\section{Correlation and \loc{5}}
\begin{figure}
\centering        \includegraphics[width=\linewidth]{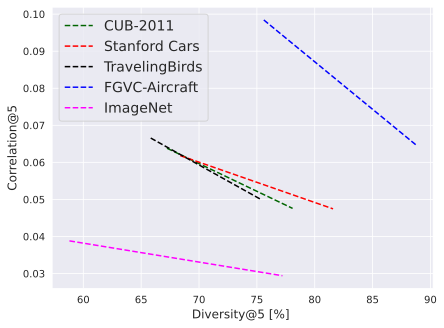}
        \caption{Relation between average absolute Pearson correlation coefficient of all top $5$ features and their \loc{5}.}
        \label{fig:corrDiv}

\end{figure}
This section discusses the relation between \loc{5} and the correlation of the used features. 
For that, the average absolute Pearson correlation coefficient $Correlation@5$ of the $5$ highest weighted features used for one class, averaged over all classes, is compared with the \loc{5} of the model.
Figure~\ref{fig:corrDiv} shows that a higher \loc{5} comes with a reduced correlation for the \gls{layerName} and \gls{NewlayerName}.
Therefore, correlation relates to \loc{5} for class-independent features. 
\section{Computational Analysis}
\label{sec:cost}
\begin{table*}[t]
\resizebox{\linewidth}{!}{
\centering
\begin{tabular}{l||c|c|c||c|c|c||c|c|c}
\toprule 
\multirow{2}{*}{Method}& \multicolumn{3}{c||}{ Accuracy  \arrowUp}& \multicolumn{3}{c||}{ T \glm{} \arrowDown} & \multicolumn{3}{c}{Total T after Dense Training \arrowDown} \\
& \cubheader{} & \stanfordheader & ImageNet & \cubheader{} & \stanfordheader & ImageNet & \cubheader{} & \stanfordheader & ImageNet \\
\midrule
\slddtable{} & 85.7$\pm$0.1 &  92.9$\pm$0.1 & 72.7$\pm$0.0 &  22&  30 & 3000 & 100 & 132 &  6600  \\
$SLDD-Model_{\mathrm{fast}}$ & 83.6$\pm$0.7 & 90.7$\pm$1.1 & - & 2 & 2 & 100 & 80 & 104 &3700   \\
\midrule
\ourtable{} &  85.9$\pm$0.4 & 92.9$\pm$0.1 & - & 22 & 30  & 3000 &    211 &282 &18000   \\
$\gls{NewlayerName}_{\mathrm{fast}}$& 85.7$\pm$0.2 & 92.9$\pm$0.1  & 74.3$\pm$0.0 &2 &2 & 100 & 131 & 170 &  6400   \\
\bottomrule
\end{tabular}
}
\caption{Impact of accelerated \glm{} on accuracy and training time consumption 
with Resnet50~[11]. Times T in minutes are rough averages or estimates using an \textit{Nvidia RTX 2080Ti}. }
\label{stab:cost}

\end{table*}

This section compares the computational cost of training~\gls{NewlayerName} with \gls{layerName} using one \textit{Nvidia RTX 2080Ti}.
There is no change for inference.
Regarding the computational cost of the iteration, both the raw training time, in our setup $3$ times $10$ epochs, and the time to fit a new sparse layer using~\glm{} is relevant. Naturally, training for $70$ instead of just $40$ epochs almost doubles the training time. However, the time \glm{} takes is reducible due to the proposed quantization and iteration, as less accuracy is required for the sparse layer. \suppl{} Table~\ref{stab:cost} shows the impact of increasing the tolerance level of the stopping criteria to $0.1$ and setting the length of the regularization paths to $5$, encoded as \enquote{fast}.
\gls{NewlayerName} allows to speed up \glm{} without losing accuracy, resulting in similar cost on ImageNet.
This setting is used for the ImageNet experiments with \gls{NewlayerName}, but we kept the time-intensive parameters for the \gls{layerName}.

\section{Global Explanations}
This section contains more exemplary images with the 5 most important features for the dense conventional \glsentryfirst{resNet} compared to our~\gls{NewlayerName} in \suppl{} Figures~\ref{app:fig:vizExamplesCardinal} to~\ref{app:fig:vizExamplesE-170}. 
The increased interpretability of the proposed~\gls{NewlayerName} is apparent due to the improved localization, measured as~\loc{5} in Section 4.2 and \suppl{} Table~\ref{stable:Diversity}, and interpretability of the features, as well as the global interpretability of each class being assigned $5$ features rather than having different weights for all $2048$ features.
The improved robustness to spurious correlations is also supported by~\suppl{} Figures~\ref{app:fig:vizExamplesCardinal} to~\ref{app:fig:vizExamplesTree_Swallow}. 
The features of \gls{NewlayerName} focus less on the spuriously correlated background, but rather on semantically meaningful attributes of the bird.

\clearpage
\begin{landscape}
\begin{table}
\resizebox{\linewidth}{!}{

\centering
\begin{tabular}{l||ccccc||ccccc||ccc|cccccc}
\toprule 
\multirow{2}{*}{Method}&  \multicolumn{5}{c||}{ Accuracy  \arrowUp}& \multicolumn{5}{c||}{ \loc{5} \arrowUp} & \multicolumn{1}{c}{No Concept} &\multicolumn{2}{c|}{Alignment $r$ \arrowUp}& \multicolumn{1}{c}{ Sparse} & \multicolumn{5}{c}{ Dependence \dependence{} \arrowDown}\\
& \cheader{} & \fheader  &    \sheader & \theader{}& \iheader & \cheader{} & \fheader  &   \sheader & \theader{} & \iheader &Supervision & \cheader{} & \theader{}&
$\gls{nperClass}\leq5$

&\cheader{} & \fheader  &    \sheader & \theader{} & \iheader \\ 
\midrule

Dense & 86.7$\pm$0.2 &  92.0$\pm$0.2 & 93.5$\pm$0.1 & 39.4$\pm$0.4 & 76.1  & 51.6$\pm$0.2 & 46.7$\pm$0.5 & 47.6$\pm$0.2 &  51.5$\pm$0.2 & 52.0 & \checkmark& 1.0$\pm$0.0 & 1.0$\pm$0.0 & \xmark & 10.4$\pm$11.0 & 24.4$\pm$0.1 & 22.2$\pm$0.8 & 22.3$\pm$0.5 & 3.2   \\ 
\midrule
\glmtable{}  & 77.7$\pm$0.4 & 89.9$\pm$0.6 & 89.1$\pm$0.5 & 36.0$\pm$0.8 & 58.0$\pm$0.0 & 46.7$\pm$0.3& 43.1$\pm$0.5 & 43.7$\pm$0.4& 47.7$\pm$0.3 & 50.0$\pm$0.0 & \checkmark& 1.0$\pm$0.0 & 1.0$\pm$0.0 & \checkmark & 49.2$\pm$0.8 & 49.3$\pm$1.5 & 49.8$\pm$0.8 & 52.7$\pm$0.8 & 53.6$\pm$0.0 \\
\gls{cbm}-joint & 82.2$\pm$0.2 &N/A & N/A&36.9$\pm$0.3  & N/A & 71.0$\pm$0.8 & N/A & N/A & 73.7$\pm$0.9  & N/A &\xmark&  3.0$\pm$0.0 &   2.8$\pm$0.0 & \xmark&  6.1$\pm$0.1 & N/A &  N/A& 5.7$\pm$0.1 &  N/A   \\
SLDD-Model & \sca$\pm$0.1 & \saa$\pm$0.2&  \textbf{\ssa}$\pm$0.5 & \sta$\pm$0.8& 72.7$\pm$0.0 & \scd$\pm$1.6 & \sad$\pm$1.8 &  \ssd$\pm$1.4 & \std$\pm$1.1 & 58.8$\pm$0.7 & \checkmark& \scal$\pm$0.1  &  \stal$\pm$0.1& \checkmark & \sct$\pm$1.2 & \sat$\pm$1.6  &\sst$\pm$0.6 & \stt$\pm$0.3 &   47.3$\pm$0.0\\  
\hdashline
Q-SENN (Ours) &\textbf{\qca}$\pm$0.4 & \textbf{\qaa}$\pm$0.2&  \textbf{\qsa}$\pm$0.1 & \textbf{\qta}$\pm$0.5& \textbf{74.3}$\pm$0.0 & \textbf{\qcd}$\pm$1.5 & \textbf{\qad}$\pm$0.8 &  \textbf{\qsd}$\pm$1.0 & \textbf{\qtd}$\pm$0.9 & \textbf{77.2}$\pm$0.6 & \checkmark& \textbf{\qcal}$\pm$0.1  &  \textbf{\qtal}$\pm$0.1& \checkmark   & \textbf{\qct}$\pm$0.4 & \textbf{\qat}$\pm$0.9  &\textbf{\qst}$\pm$0.7 & \textbf{\qtt}$\pm$0.5 &   \textbf{30.7}$\pm$0.0\\
\bottomrule
\end{tabular}
}
\caption{Comparison across all desired metrics with~\resnet{}. Best results among comparable interpretable models  are in bold.
\textit{N/A} indicates inapplicability due to missing annotations, 
while values with \xmark{} cannot be reasonably compared to the remaining rows due to sparsity or supervision.
Accuracy, \loc{5} and Dependence~\dependence{} are measured \inpercent{}. Note, that our proposed \gls{NewlayerName} maintains most or all of the accuracy of dense models while simultaneously heavily improving the interpretability.}
\label{stab:overview}
\end{table}
\begin{table}
\resizebox{\linewidth}{!}{
\centering
\begin{tabular}{l||cccc||cccc||cc|cccc|cccc}
\toprule 
\multirow{2}{*}{Method}& \multicolumn{4}{c||}{ Accuracy  \arrowUp}& \multicolumn{4}{c||}{ \loc{5} \arrowUp} & \multicolumn{2}{c|}{Alignment $r$ \arrowUp} & \multicolumn{4}{c|}{ Dependence \dependence{} \arrowDown} & \multicolumn{4}{c}{Binary Features \arrowUp}\\
& \cheader{} & \fheader  &    \sheader & \theader{}&  \cheader{} & \fheader  &   \sheader & \theader{} &  \cheader{} & \theader{}&
\cheader{} & \fheader  &    \sheader & \theader{} &\cheader{} & \fheader  &    \sheader & \theader{}  \\ 
\midrule
Q-SENN (Ours)  & \textbf{85.9}$\pm$0.4 & \textbf{92.1}$\pm$0.2 & {92.9}$\pm$0.1 & {67.3}$\pm$0.5 & \textbf{78.1}$\pm$1.5 & \textbf{88.7}$\pm$0.8 & \textbf{81.6}$\pm$1.0 & \textbf{75.4}$\pm$0.9 & {2.2}$\pm$0.1 & {2.1}$\pm$0.1 & {30.0}$\pm$0.4 & \textbf{30.5}$\pm$0.9 & {29.3}$\pm$0.7 & {30.7}$\pm$0.5 & {80.4}$\pm$8.7 & {71.2}$\pm$8.1 & {90.0}$\pm$4.4 & \textbf{84.0}$\pm$2.5 \\
 w/o Quantization & {85.4}$\pm$0.3 & {91.7}$\pm$0.2 & {92.6}$\pm$0.1 & \textbf{70.0}$\pm$0.7 & {64.4}$\pm$1.7 & {63.3}$\pm$2.0 & {65.3}$\pm$2.7 & {61.8}$\pm$1.3 & \textbf{3.1}$\pm$0.1 & \textbf{2.8}$\pm$0.1 & {46.8}$\pm$0.6 & {55.6}$\pm$1.4 & {47.7}$\pm$0.6 & {47.3}$\pm$0.4 & {53.2}$\pm$4.7 & {34.8}$\pm$4.8 & {59.2}$\pm$9.8 & {70.8}$\pm$3.7  \\
w/o Iteration & \textbf{85.9}$\pm$0.4 & {92.0}$\pm$0.2 & \textbf{93.0}$\pm$0.1 & {63.9}$\pm$0.6 & {70.9}$\pm$1.7 & {85.2}$\pm$1.1 & {73.5}$\pm$1.7 & {68.9}$\pm$1.0 & {1.7}$\pm$0.0 & {1.5}$\pm$0.1 & \textbf{29.2}$\pm$0.4 & {30.6}$\pm$0.8 & \textbf{28.6}$\pm$0.5 & \textbf{30.5}$\pm$0.6 & \textbf{90.4}$\pm$2.9 & \textbf{88.0}$\pm$4.2 & \textbf{96.4}$\pm$2.7 & {78.4}$\pm$5.0 \\
\bottomrule
\end{tabular}
}
\caption{Ablation Study on Impact of iteration and quantization. Binary Features are the percentage of the features for which mean shift applied on the training distribution returns exactly 2 clusters.}
\label{stab:abl}
\end{table}
\end{landscape}
\comparison{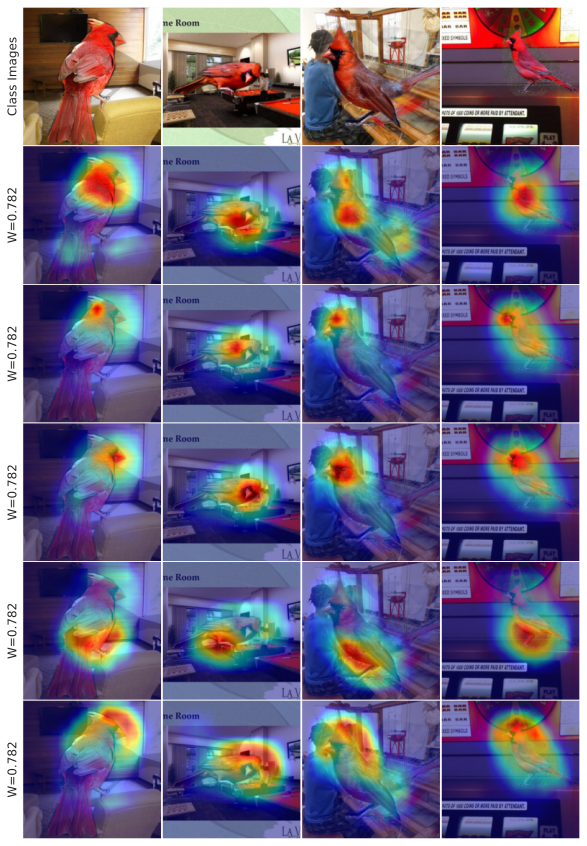}
\comparison{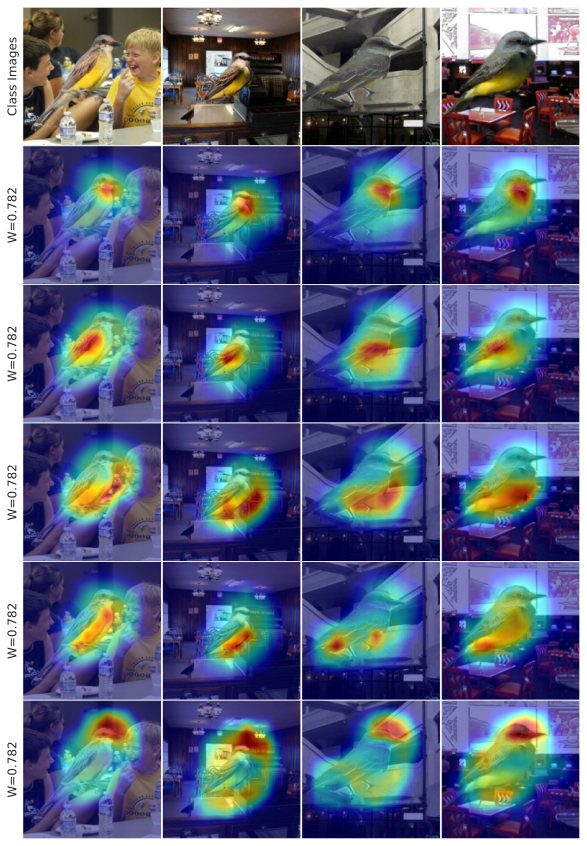}
\comparison{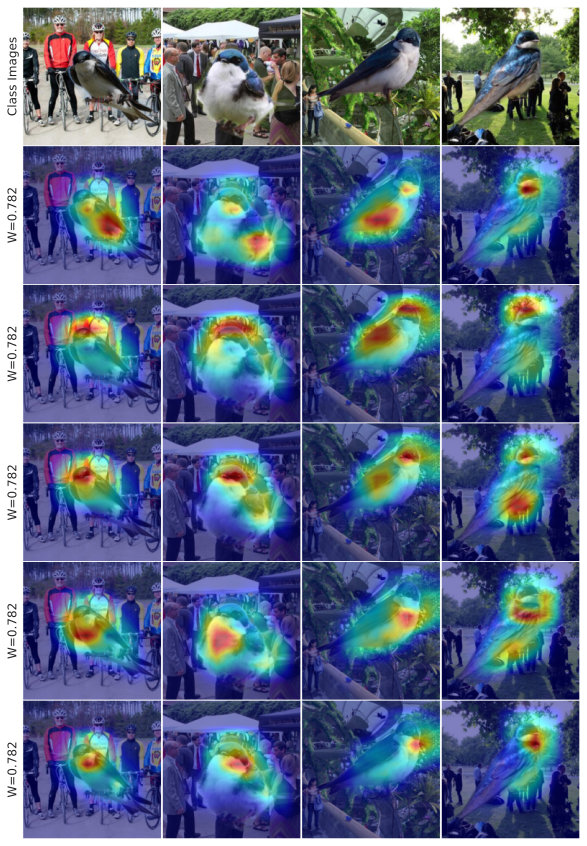}
\comparison{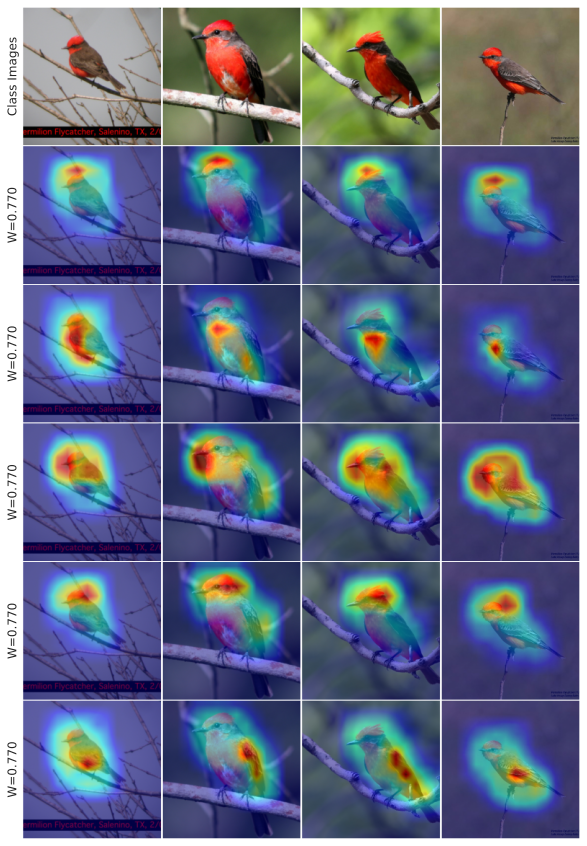}
\comparison{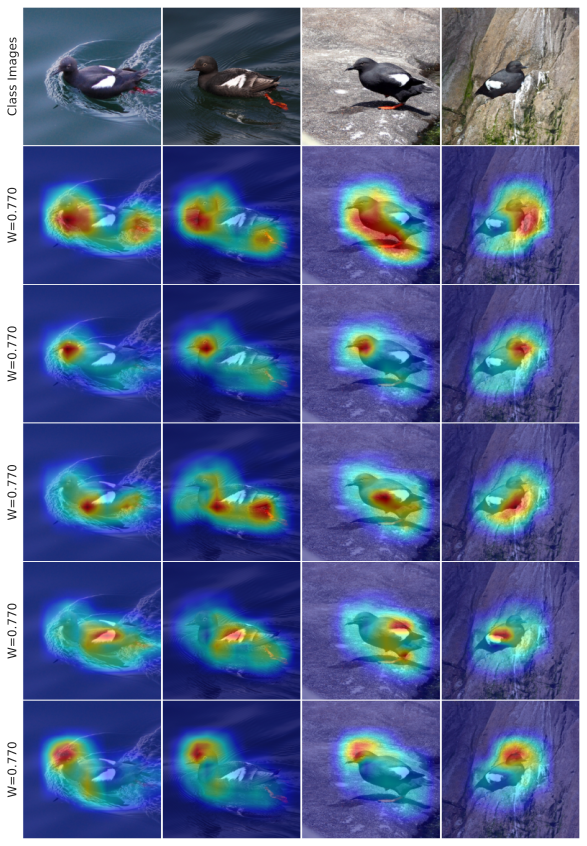}
\comparison{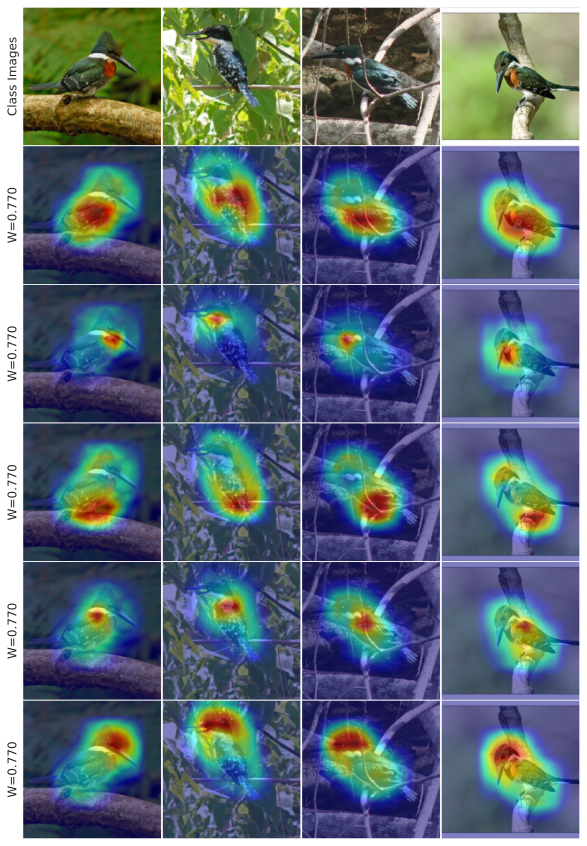}
\comparison{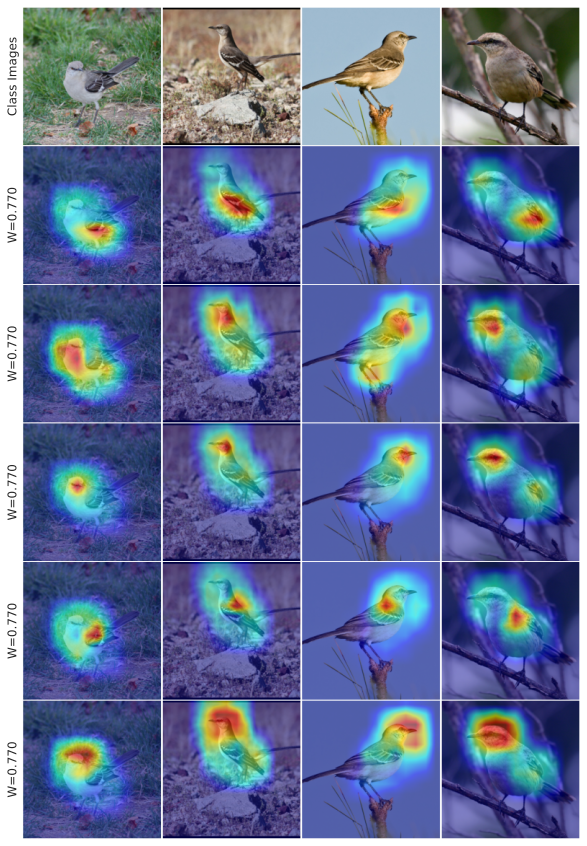}
\comparison{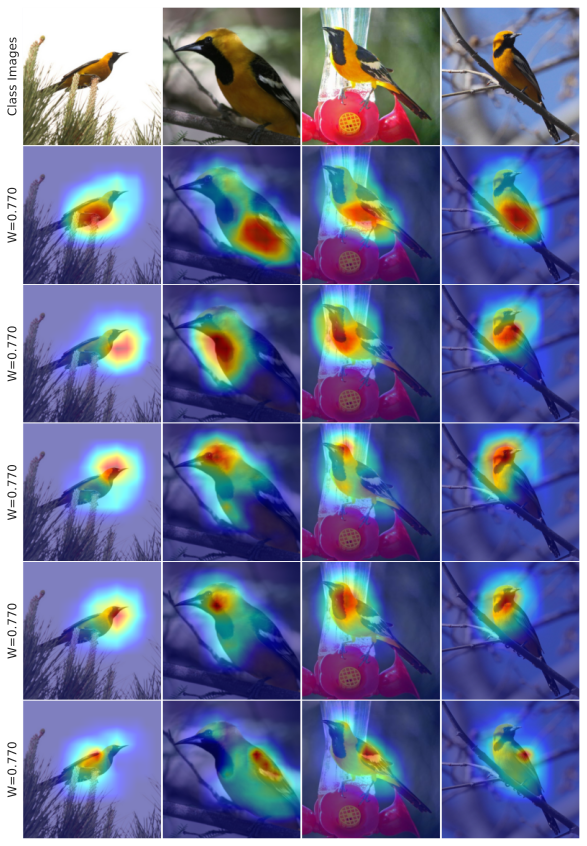}
\comparison{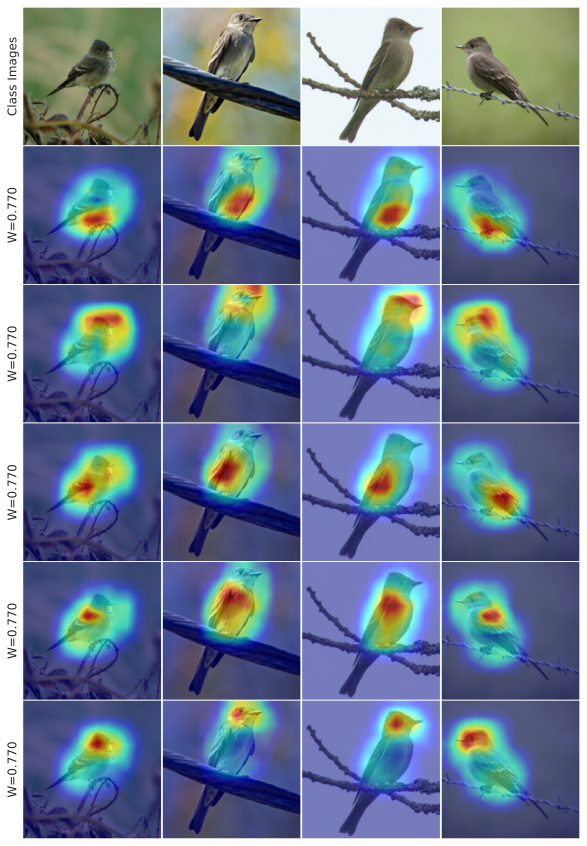}
\comparison{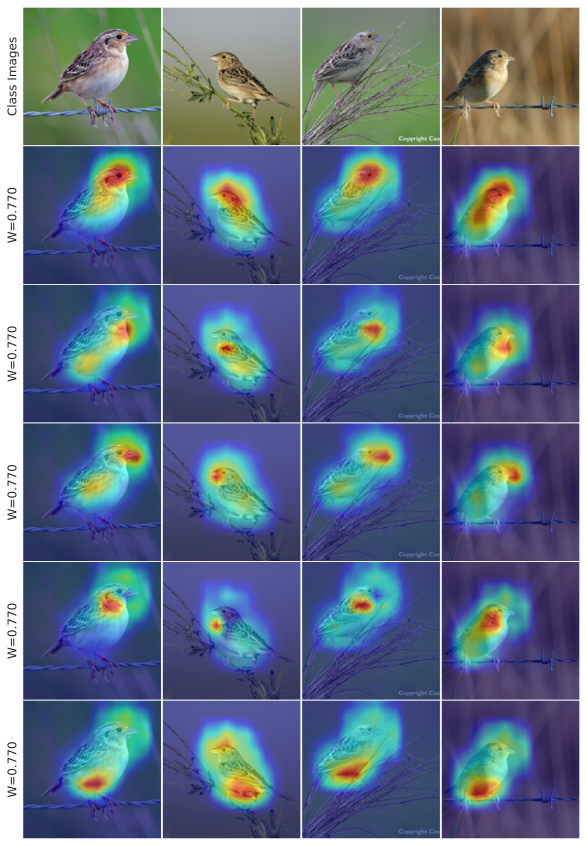}
\comparison{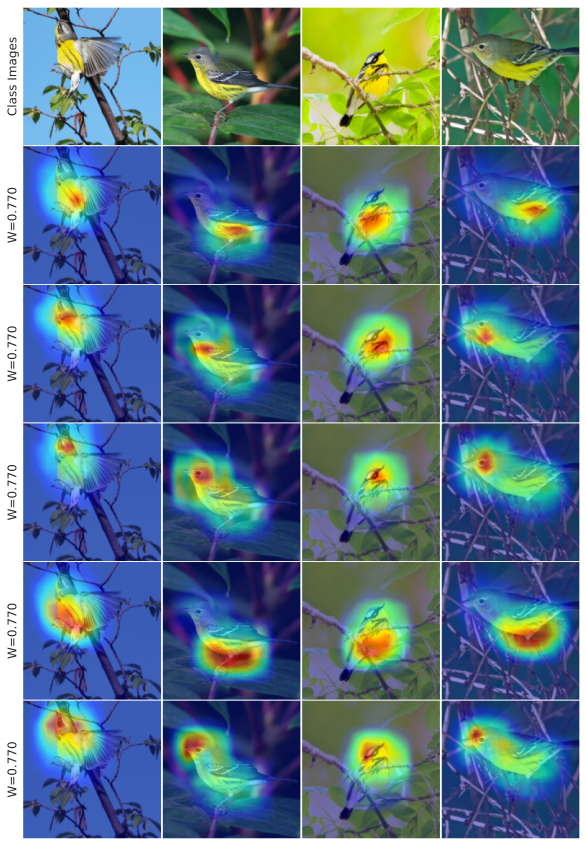}
\comparison{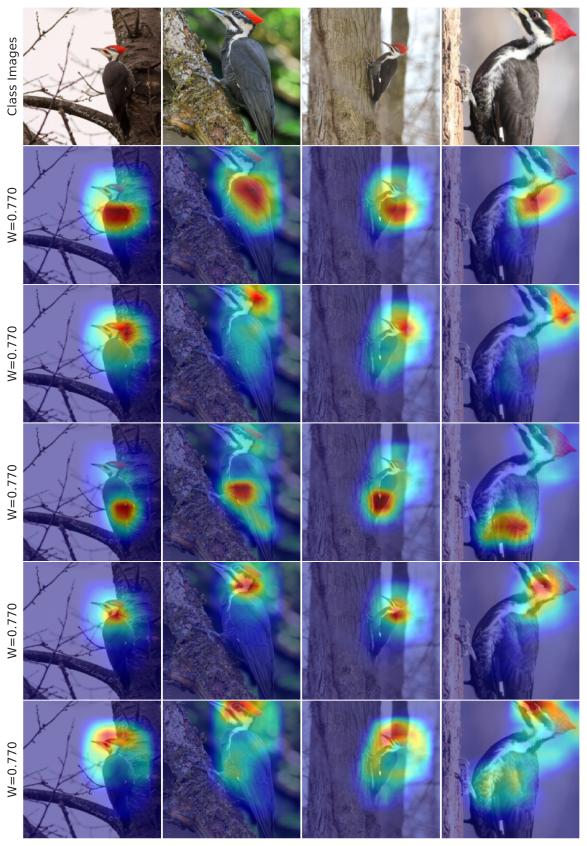}
\comparison{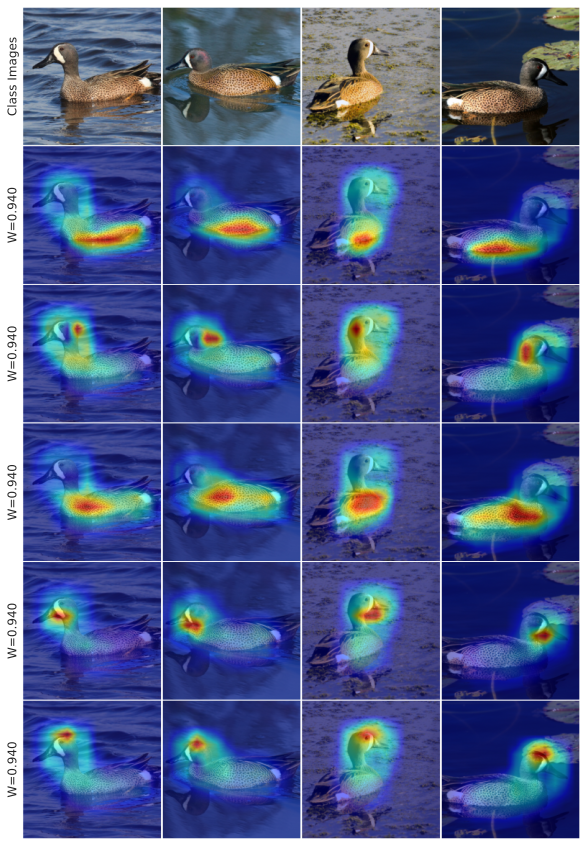}
\comparison{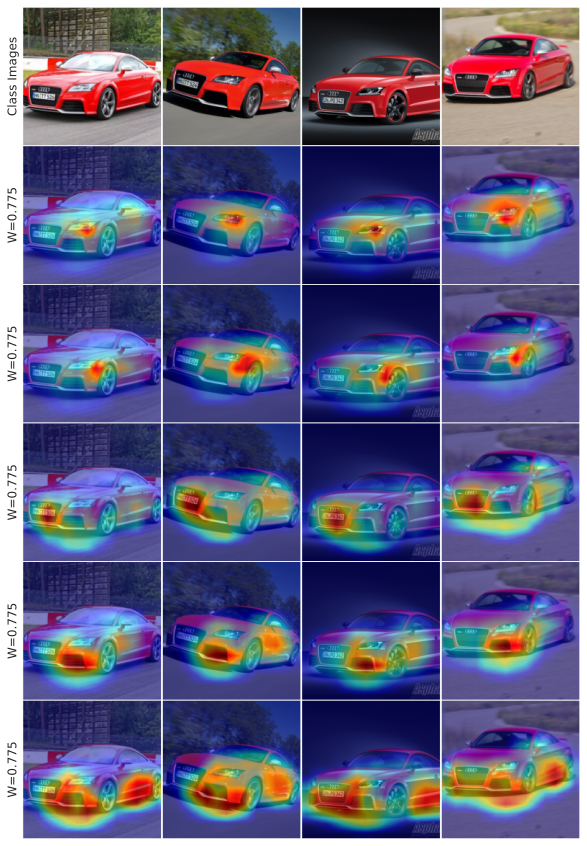}
\comparison{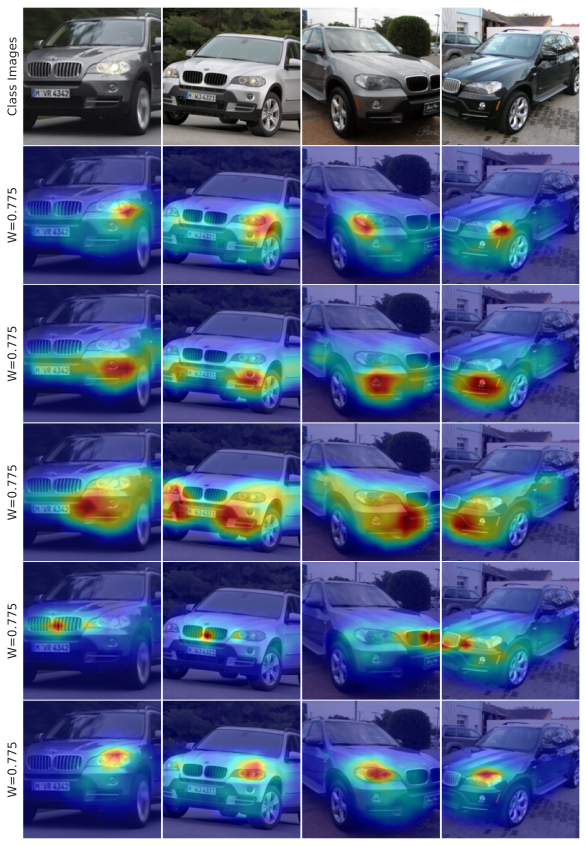}
\comparison{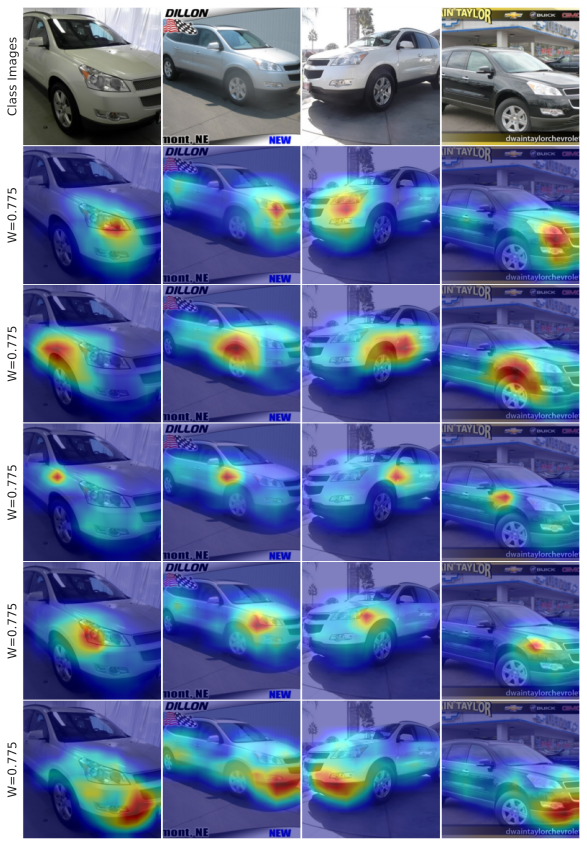}
\comparison{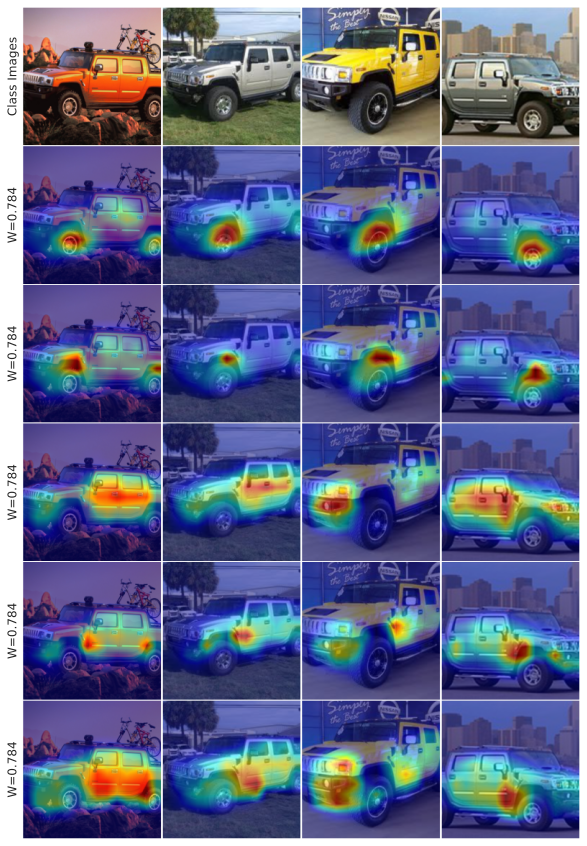}
\comparison{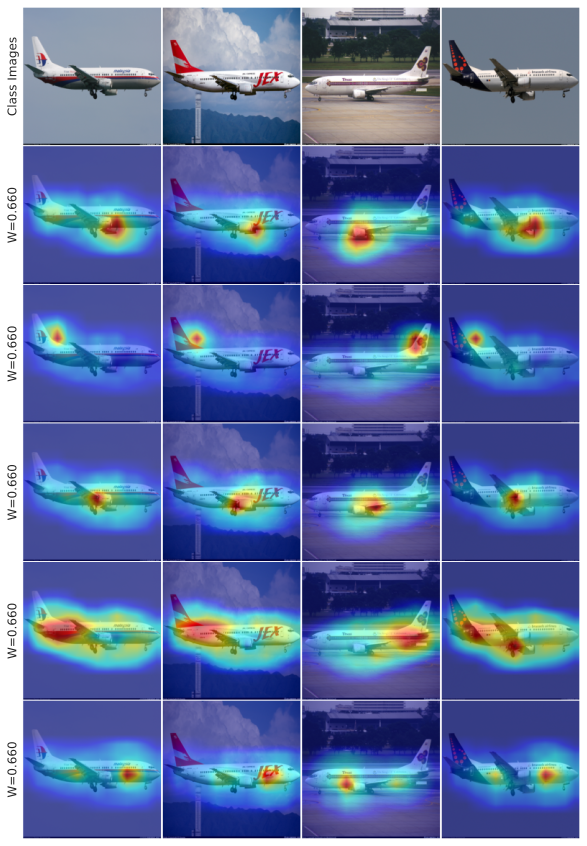}
\comparison{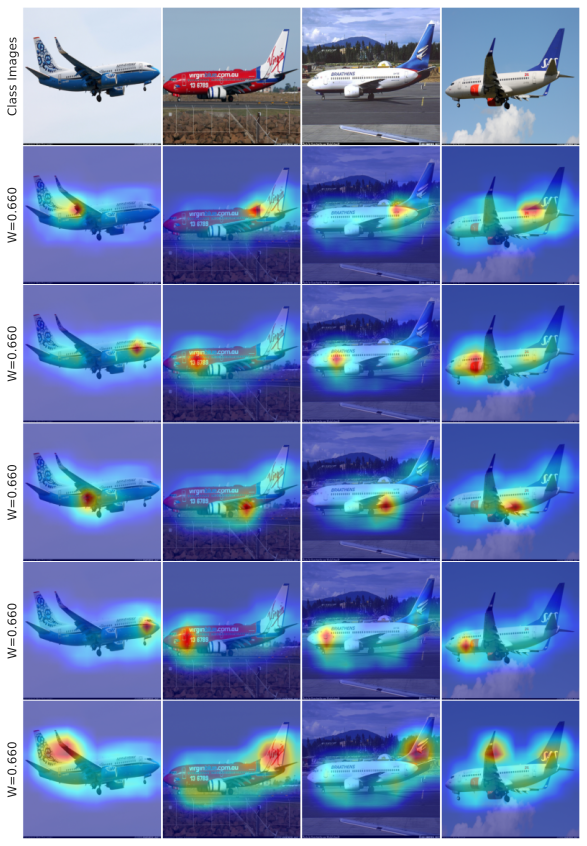}
\comparison{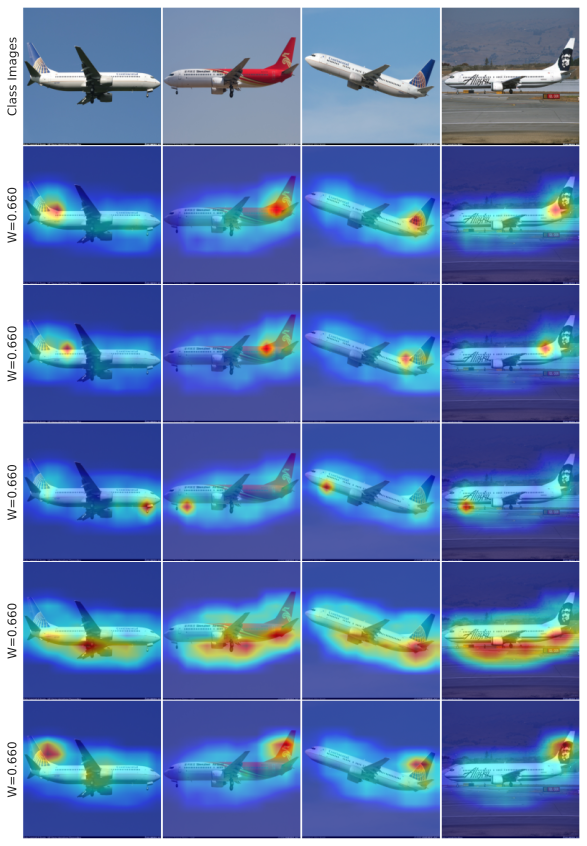}
\comparison{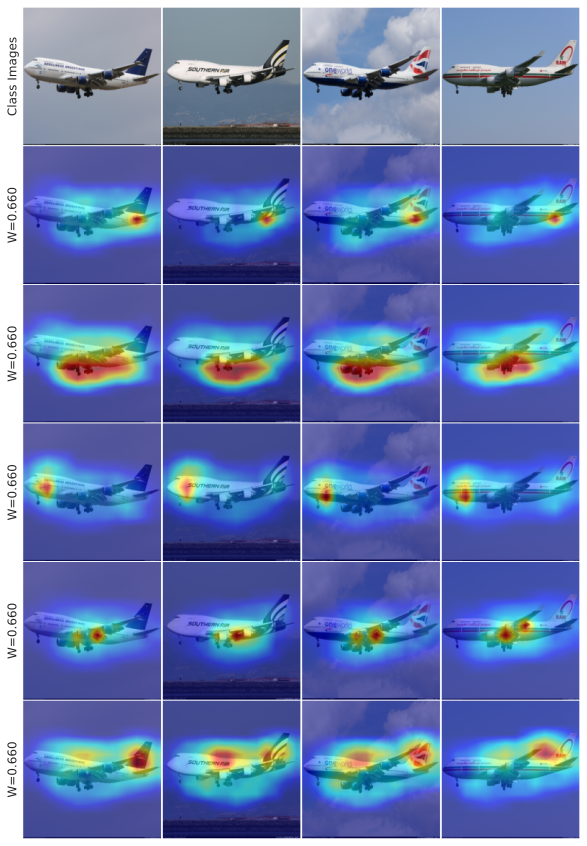}
\comparison{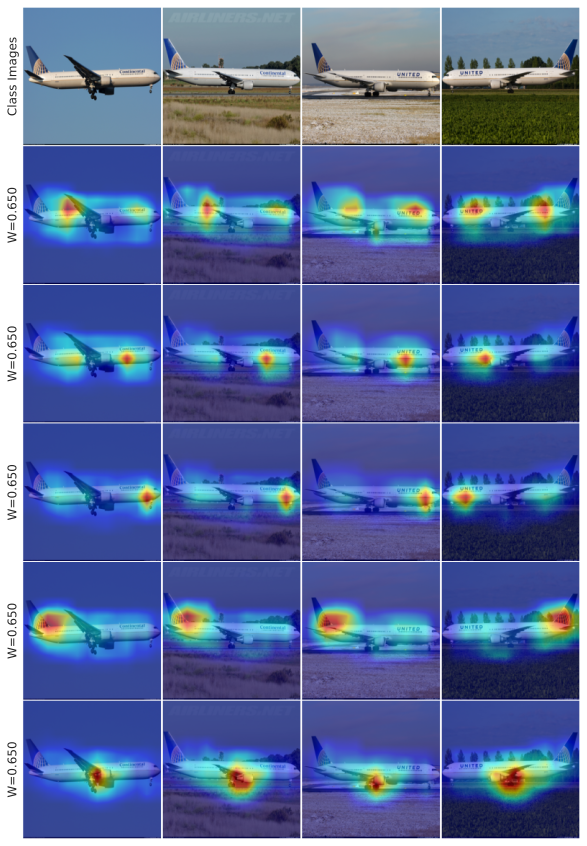}
\comparison{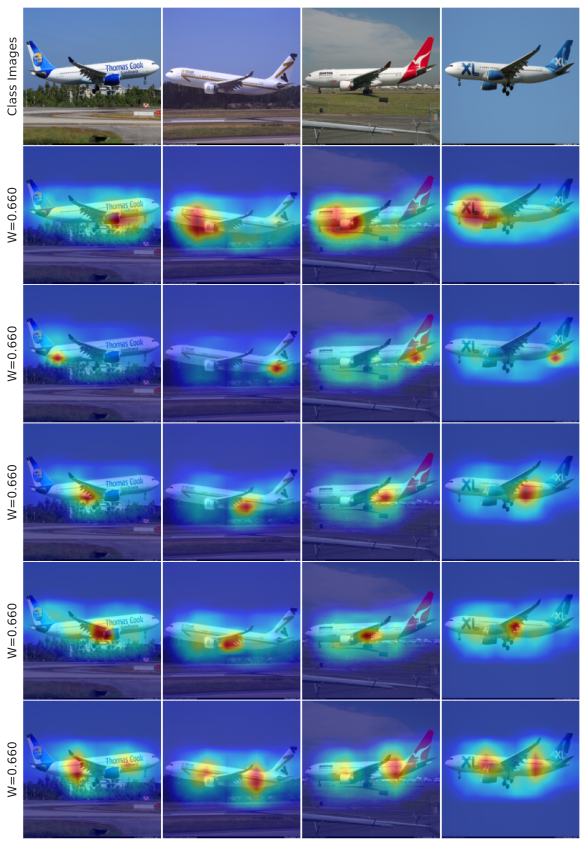}
\comparison{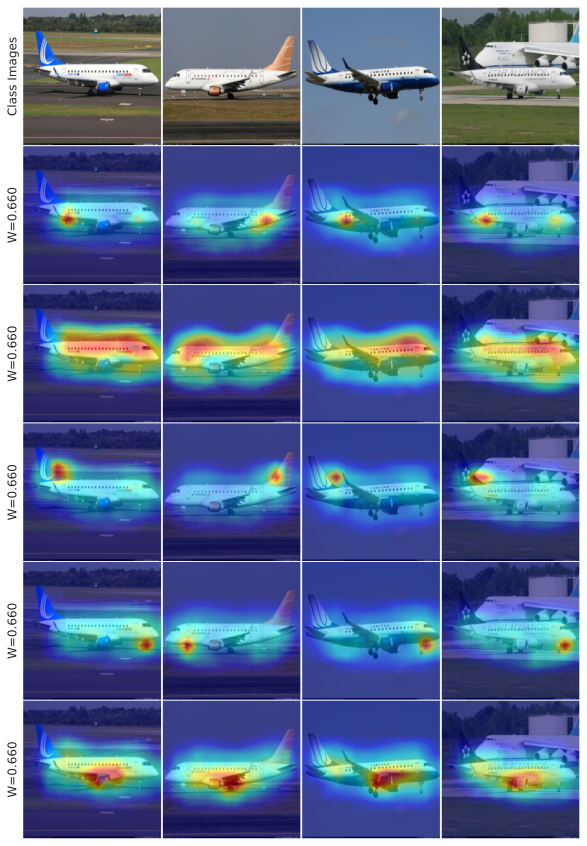}
\clearpage


\end{document}